\documentclass[10pt,twocolumn,letterpaper]{article}

\usepackage{iccv}
\usepackage{times}
\usepackage{epsfig}
\usepackage{graphicx}
\usepackage{amsmath}
\usepackage{amssymb}
\usepackage{algorithm, algorithmic}
\usepackage{booktabs}
\usepackage{placeins}
\usepackage{multirow}

\def\x{{\bf{x}}}

\def\z{{\bf{z}}}
\def\w{{\bf{w}}}
\def\p{{\bf{p}}}
\def\R{{\mathbb{R}}}
\def\D{{\mathcal{D}}}
\def\B{{\mathcal{B}}}

\def\U{{\mathcal{U}}}

\usepackage[pagebackref=true,breaklinks=true,letterpaper=true,colorlinks,bookmarks=false]{hyperref}

\iccvfinalcopy

\begin{document}

\title{Improving Identity-Robustness for Face Models}

\author{Qi Qi\thanks{Most of the work was done while the author was an intern at Netflix. The first version of the paper was released at April 7th, 2023.}\\
The University of Iowa\\
Iowa City, IA\\
{\tt\small qi-qi@uiowa.edu}
\and
Shervin Ardeshir\\
Netflix\\
Los Gatos, CA\\
{\tt\small shervina@netflix.com}
}

\maketitle

\begin{abstract}
Despite the success of deep-learning models in many tasks, there have been concerns about such models learning shortcuts, and their lack of robustness to irrelevant confounders. When it comes to models directly trained on human faces, a sensitive confounder is that of human identities. Many face-related tasks should ideally be identity-independent, and perform uniformly across different individuals (i.e. be fair). One way to measure and enforce such robustness and performance uniformity is through enforcing it during training, assuming identity-related information is available at scale.
However, due to privacy concerns and also the cost of collecting such information, this is often not the case, and most face datasets simply contain input images and their corresponding task-related labels. Thus, improving identity-related robustness without the need for such  annotations is of great importance. Here, we explore using off-the-shelf face-recognition embedding vectors, as proxies for identities, to enforce such robustness. We propose to use the structure in the face-recognition embedding space, to implicitly emphasize rare samples within each class. We do so by weighting samples according to their conditional inverse density (CID) in the proxy embedding space. Our experiments suggest that such a simple sample weighting scheme, not only improves the training robustness, it often improves the overall performance as a result of such robustness. We also show that employing such constraints during training results in models that are significantly less sensitive to different levels of bias in the dataset.
\end{abstract}

\vspace{-0.05in}
\section{Introduction}
\vspace{-0.05in}

\begin{figure*}[t]
\centering
\includegraphics[width = \linewidth]{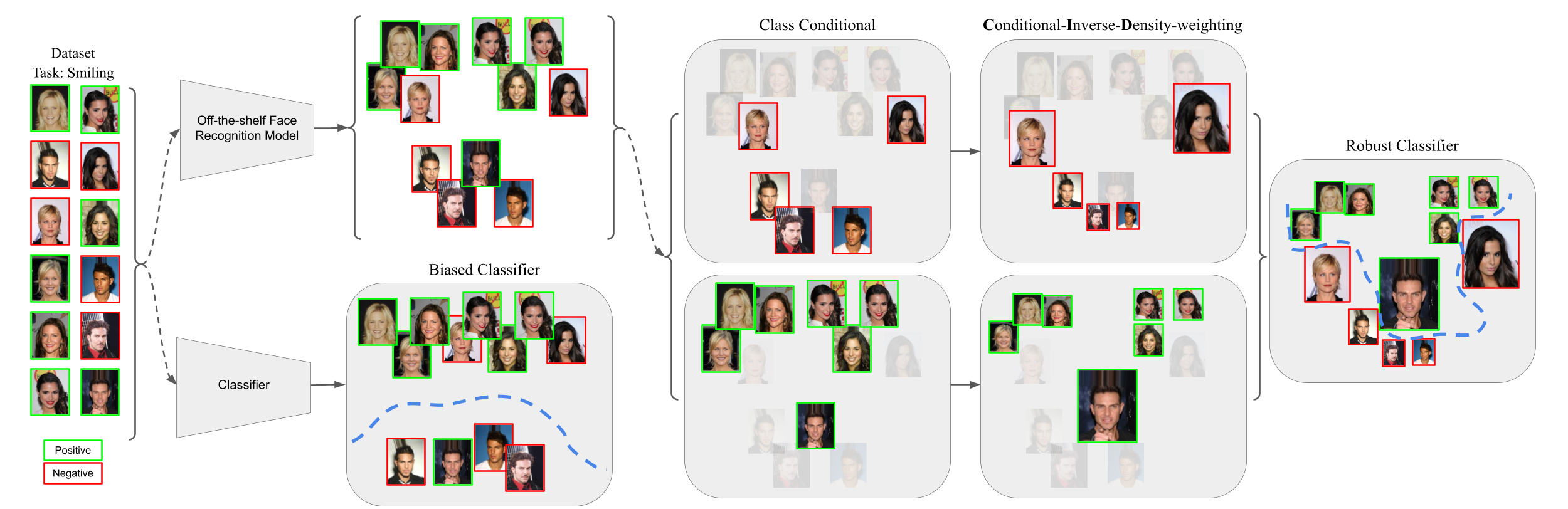}
\caption{Toy example visualizing our proposed approach. The task is predicting if a face image is smiling (green) or not (red). The Biased Classifier shows how a biased dataset could lead to a model latching on to spurious features (identity) for an identity-independent task (smiling). We propose extracting face-recognition embeddings and using the structure in that space to weight rare samples within each class. More specifically, for each class (green or red), each sample is weighted based on its class-conditioned inverse density in the proxy (face recognition) embedding space. As a result, in each class, the rare samples are emphasized in the Robust Classifier.}
\label{fig:face_model_illstration}
\end{figure*}

Given the success of machine learning models, and their deployment at scale, having a more extensive evaluation of the robustness of such models is of utmost importance. Given the nature of training such models, there is always the potential for these models to rely on irrelevant and spurious shortcuts. Relying on such shortcuts could have immense negative consequences when the dataset and tasks are defined around humans. A prevalent type of such datasets and tasks are those defined on human faces, ranging from regression tasks such as estimating pose\cite{albiero2021img2pose}, facial-landmarks\cite{wu2019facial}, etc, to classification tasks such as facial-expressions classification\cite{huang2019facial}, and generative tasks such as avatar creation\cite{alldieck2018detailed}, etc. A common attribute of many of such face-centric tasks is the fact that the model performance, should be identity independent by definition, yet this aspect of a model is often not taken into account during training and evaluation. Two models trained on a face-related task can have similar overall performance, but very different levels of robustness across different individuals. The toy example in Figure~\ref{fig:face_model_illstration} illustrates this concept. This disparity in performance often gets baked into the model due to bias in the training data, as data points belonging to different subpopulations may have a different level of class imbalance. More specifically, if person 1 smiles 90\% of the time, and person 2 smiles 10\% of the time, a smile classifier can easily latch on to the facial features of person 1 as a shortcut to reduce training loss significantly. Thus, it could always label images of person-1 as smiling, because of the person's identity, and not the facial expression. 

Awareness of identity/group labels would allow for mitigation approaches to prevent such bias, such as recent efforts in adversarial training~\cite{zhang2018mitigating, elazar2018adversarial}, model interpretation method \cite{rieger2020interpretations} and objective regularization~\cite{bechavod2017penalizing}, which aim to reduce the disparity between different groups using the ground-truth group labels $g\in G$. In many practical scenarios, however, such information is not available at scale during training and evaluation. Also, collecting such detailed annotation could be costly and undesirable due to three main reasons: First, annotating every sample data point with all their potential types of group-membership information could be extremely costly. Second, collecting and maintaining such detailed categorical labels on human faces raise data-privacy concerns. And third, the nature of many types of such group memberships may be extremely subjective. In addition to the previous hurdles in obtaining such data, most current large-scale datasets, lack such annotations at scale, which is another testament to the need for approaches that are not reliant on the availability of such additional information. As a result, improving fairness when the ground-truth group labels $g\in G$ are unknown is of utmost importance, and has given rise to an area of research often referred to as "fairness under unawareness". When it comes to "fairness under unawareness" for face models, the only earlier work is ~\cite{ardeshir2022estimating} which aims to measure the performance disparity of a model in the absence of group information. A disparity method (Disparity across Embedding Neighborhoods) is proposed, which approximates Rawlsian Max-Min (RMM) across groups $g\in G$, solely based on face-recognition embedding vectors. The neighbors of a sample are defined as the samples whose euclidean distance in the face-recognition embedding space is less than a predefined threshold. The aforementioned work solely focused on approximating disparity for a given model. In this work, however, we focus on using such intuition to reduce such disparity during training and directly optimize for such an objective. In other words, given a face dataset and solely its task labels, and without any group information, we explore if we can use embeddings from an off-the-shelf face recognition model to reduce the performance disparity of such a model across different individuals. 

As identity bias is often induced due to class imbalance within a subpopulation of a dataset, we propose to exploit the structure of the training data points in the face-recognition embedding space and enforce class balance for any subpopulation that shares similar facial features. This is achieved by weighting the samples according to their conditional inverse density (CID) in the proxy embedding space, resulting in equalizing the effect of task-positive and task-negative labels for each local neighborhood in the proxy embedding space. In other words, we aim to equalize the total weight for positive and negative samples for people whose facial features look like any arbitrary embedding $z$. Our experiments show that such a simple sample weighting scheme not only reduces the performance disparity of the trained model across different individuals and groups but also makes the training more robust to distribution shifts between train and test, and different levels of dataset bias. We evaluate such robustness by designing a stress test where we artificially manipulate the bias in the dataset, and control its level of bias.



\vspace{-0.1in}
\section{Related Work}

\noindent
The bias mitigation methods can be broadly categorized into two groups based on the availability of group information, denoted by $G$. When $G$ is available, the proposed methods usually explicitly incorporate such group information into the training process to reduce bias, such as penalizing the group difference as a regularizer~\cite{bechavod2017penalizing,beutel2019fairness}, enhancing fair representations via contrastive learning only using task-relevant features by constructing hard negative pairs from different groups~\cite{zhangfairness, park2022fair, du2021fairness}, learning an adversary head to reduce models' ability to distinguish group-relevant features that amplify biases~\cite{zhang2018mitigating, wadsworth2018achieving}, and so on.

Our work can be categorized under the "fairness under unawareness" umbrella, where group information $G$ is unavailable during training. Due to the lack of group information, improving model robustness to minimize spurious correlations has become one of the mainstream methods to enhance model fairness. This can be achieved through various methods such as invariant risk minimization~\cite{arjovsky2019invariant,adragna2020fairness},  distributionally robust optimization~\cite{ben2013robust, caton2020fairness, sagawa2019distributionally,hashimoto2018fairness,li2021tilted, qi2020attentional, lahoti2020fairness}, and class balancing methods~\cite{yan2020fair,cui2019class,huang2016learning, wang2017learning}. 

\noindent
\textbf{Invariant Risk Minimization (IRM)}
IRM~\cite{arjovsky2019invariant} was proposed to address the domain-shifting problems by learning invariant feature representations that have good generalization ability between training and testing. To achieve this,
IRM optimizes the losses across different data distributions to improve model robustness.
~\cite{adragna2020fairness} shows the effectiveness of IRM in reducing the disparity between different protected groups in the toxicity classification task on Civil Comments natural language process dataset when $G$ is unavailable.

\noindent
\textbf{Distributionally Robust Optimization (DRO)} Compared with ERM which minimizes the average sample losses, DRO aims to focus on the largest errors~\cite{ben2013robust, caton2020fairness} by assigning robust weights $p_i$ to samples proportional to the loss scales.
To verify the validity of DRO in improving fairness under unawareness,
~\cite{liu2021just} empirically shows that most of the samples with the largest errors belong to the worst group, ~\cite{hashimoto2018fairness} theoretically proves that DRO can control group disparity amplification in every iteration, and~\cite{li2021tilted, qi2020attentional} shows the effectiveness of DRO in different fair applications.
Recently, ~\cite{lahoti2020fairness} proposed a generalized DRO method, namely Adversarially Reweighted Learning (ARL), by parameterizing the robust weight $p_i$ with an adversarial network $\phi$ and leveraging the concept of a computationally-identifiable subgroup of largest errors~\cite{hebert2018multicalibration} to improve model fairness.

\noindent
\textbf{Class Balancing}~\cite{yan2020fair} proposed  a cluster-based balancing method by generating the minority samples for each cluster using the upsampling K-Means SMOTE~\cite{last1711oversampling}, however, this upsampling clustering-based method only applicable to small tabular data and lead to excessive training time for larger datasets. Alternatively,
class balancing reweighting methods~\cite{cui2019class,huang2016learning, wang2017learning} are widely used to improve model robustness in large datasets by assigning weights to balance the contribution of different classes. These weights of each sample are typically inversely proportional to the number of class samples, which helps to improve the performance of minority classes and alleviates the spurious correlations between the sensitive groups and classes incurred due to the lack of samples in minority class.

To summarize, the IRM, DRO, and class-balancing methods rely on the implicit preservation of sensitive group (categorical) information in model predictions, either through losses~\cite{adragna2020fairness, ben2013robust, caton2020fairness, sagawa2019distributionally,hashimoto2018fairness,li2021tilted, qi2020attentional}  or feature representation embeddings~\cite{lahoti2020fairness,yan2020fair}. However,  in our case, this assumption does not hold as we are focusing on face-centric tasks that are defined identity independent. Therefore, our CID method resorts to face recognition embeddings to have better group proxies. In the supplementary, we provide comparisons and draw parallels between our approach and the aforementioned methods under certain assumptions.

\section{Approach}
\label{sec:approach}
Given a dataset of images of faces, and an identity-independent face-related task such as predicting a facial expression (e.g. smiling), we aim to train a classifier that performs robustly across face images of different people. We refer to training labels related to the task of interest (smiling), as \textit{task labels}. We assume that such labeling (whether a face is smiling or not) is given to us for training and test set. On the contrary, we assume that no \textit{identity} label is given to us during training. Identity labels specify which images belong to which person (person-1, person-2, ...), across which we would like to enforce fairness/robustness.  
We also assume that we have access to an off-the-shelf face-recognition model, using which we can extract an embedding for each face image. Our goal is to train a model for that task, that performs robustly (fairly) across different individuals on the test set. Please note that in our experiments, we solely use the identity-labeled test sets to validate the robustness of our approach, and we do not use such labels during training. Formally, given a dataset $\D = \{X\times Y\} = \{(\x_i,y_i )\}_{i=1}^{|\D|}$ with size $n=|D|$, the total number of classes $C$, i.e, $|Y| = C$. $\D_{y} = \{(\x_i,y_i)|y_i = y, i\in [1,\cdots,|\D|]\}$ represents the samples whose task label is $y\in Y$. $g_i\in G$ denotes the identity/group that sample $i$ belongs to, across which performance disparity should be mitigated. Under our setup, group/identity labels $G$ are unavailable during training. Instead, the embedding vectors $\{\z_i\}_{i=1}^{|\D|}$ are extracted from a face recognition model and are provided as proxies for the group/identity membership.


\vspace{-0.05in}
\section{Training}
\vspace{-0.05in}
\label{sec:training}
Inspired by recent efforts in ~\cite{diana2021minimax,hashimoto2018fairness,lahoti2020fairness}, we define our objective as a min-max form, which encourages emphasis the performance of the model on the least accurate areas of the embedding space:
 \vspace{-0.05in}
\begin{align}
\label{eqn:obj}
\min\limits_{\w}  \sum\limits_{i=1}^n&  \frac{p_i^\tau}{Z_{y_i}}\ell(\w;\x_i,y_i)  \\
\label{eqn:robust_fairness_constraint}
\text{s.t} \arg\max\limits_{\p_i\in\Delta_{\D_{y_i}}}  \sum\limits_{j\in \D_{y_i}} &p_{ij}\z_i^\top\z_j - \tau \text{KL}( \p_i, \frac{\textbf{1}}{|\D_{y_i}|})
\vspace{-0.2in}
\end{align}
where $p^\tau_i:= p_{ii}$ denotes the sample weight, $\ell(\w;\x_i,y_i)$ denotes the prediction loss, and $Z_{y_i} = \sum_{i\in \D_{y_i}}p^{\tau}_{i}$ is the class-level normalization parameter to guarantee each class contributes equally. To obtain $p_{i}^\tau$, the maximum constraint in (\ref{eqn:robust_fairness_constraint}) is imposed on the pairwise similarity of proxy embedding vectors, leveraging the proxy neighborhood structure associated with each sample. To be more specific, for $\forall (\x_i,y_i)\sim \D$, $\p_i= (p_{i1}, \cdots,p_{ii},\cdots, p_{i|\D_{y_i}|})$ refers to the weight assigned to each sample based on $\{\z_i^\top\z_j \}_{j\in\D_{y_i}}$ and satisfies $\Delta_{\D_{y_i}}:=\{\sum_{j} p_{ij} = 1, p_{ij}\geq 0\}$. The KL divergence regularizer  $\sum_jp_{ij}\log(|\D_{y_i}|p_{ij})$ between the uniform distribution $1/|\D_{y_i}|$ and the pairwise weights $\p_i$
encourages the model to focus on the local neighborhood. The regularizer hyperparameter $\tau$ measures the proximity and magnitude of the neighborhood, which will be explained in next section.


 

\subsection{Batch-wise Implementation using Conditional Inverse Density (CID)}
\label{sec:rpse}
Here we explain how we practically optimize the objective mentioned above using a sample-weighting scheme based on the conditional inverse density (CID for short) of each datapoint in the proxy embedding space.
To expand, we consider the practical batch-wise training scheme such that the constraint set $\D_{y_i}$ in Eqn~(\ref{eqn:robust_fairness_constraint}) is defined as the samples having the same task labels in the current batch $\B$, i.e, $\B_{y_i}$ (thus, conditioned on task label).  
Thanks to the strong concavity of $\p_i$ in~(\ref{eqn:robust_fairness_constraint}) and the specific structure of KL divergence, the close form solution of $p_i^{\tau}:= p_{ii}$ is obtained by taking the first derivative of $\p$ in (\ref{eqn:robust_fairness_constraint})  equals to 0, i.e,
\vspace{-0.1in}
\begin{align}
\label{eqn:p_ii}
     p_{i}^\tau =  \frac{\exp(\frac{\z_i^\top\z_i}{\tau})}{ \sum\limits_{k=1}^{|\B_{y_i}|}\exp(\frac{\z_i^\top\z_k}{\tau})}
\end{align}
\vspace{-0.1in}

where the numerator is the exponential of the inner product of the proxy embedding vector $\z_i$ of sample $(\x_i, y_i)$.  The denominator explores the neighborhood 
proxy structure by aggregating of the exponential pairwise similarities of proxy vectors between sample $(\x_i, y_i)$ and $\B_{y_i}$. 
Even though the constraint set is defined in $\B_{y_i}$, the skewness property of exponential function $\exp(\cdot/\tau)$ for  large similarities pairs encourages the denominator to focus on the local neighbors of $(\x_i, y_i)$ that share the same facial features. 
$p_i^{\tau}\in (0, 1]$ represents the importance of the sample $(\x_i, y_i)$ in the local neighborhood. The fewer the samples in the local neighborhood, the higher the $p_i^\tau$. Hence, $p_i^\tau$ is inverse proportional to the class-conditional sample density in the local neighborhood and emphasizes on the rare samples within each class.

\begin{figure}[t]
    \centering
\includegraphics[width = \linewidth]{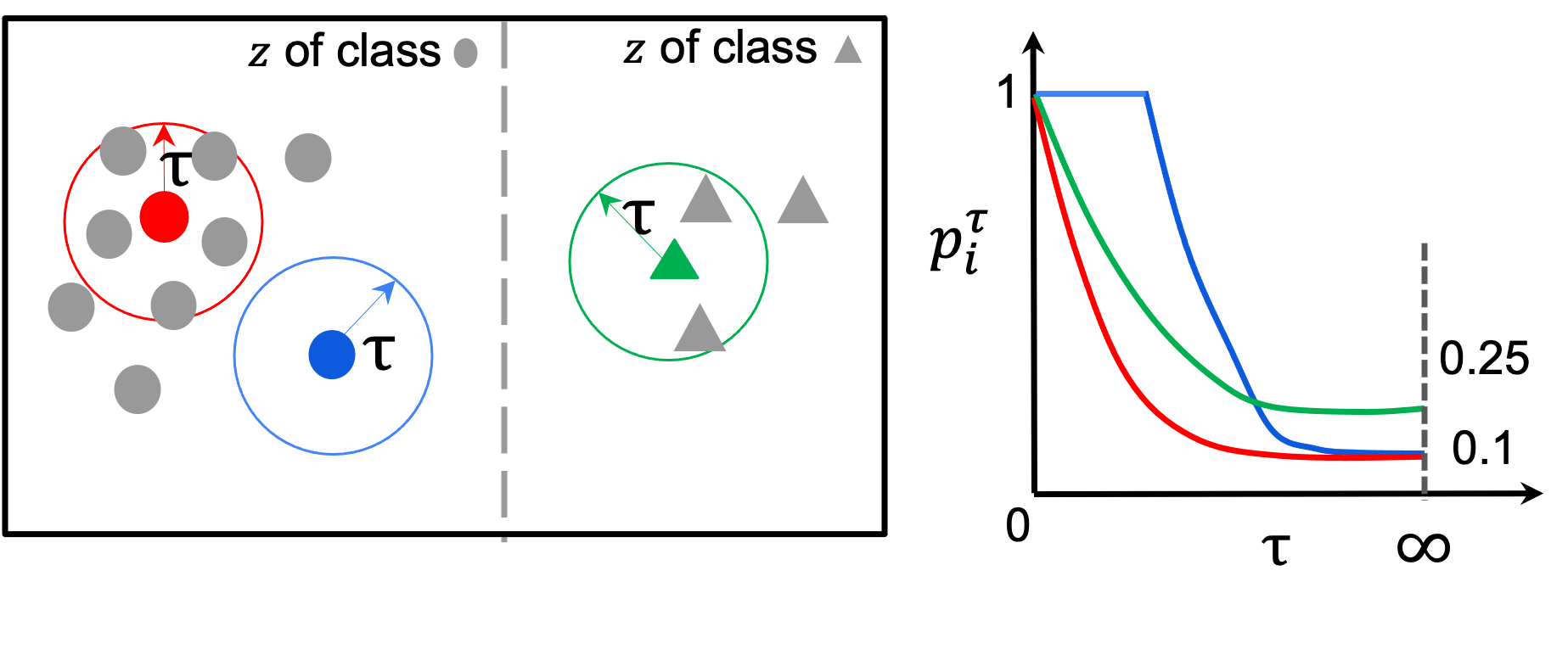}
\vspace{-0.2in}
    \caption{Visualizing the effect of $\tau$ on $p_i^\tau$ and the soft local neighborhood in the proxy vector space $\{\z\}_{i=1}^{|\B|}$ and $\z\in \R^2$. We plot the $p_i^
    \tau$ curves calculated according to equation~(\ref{eqn:p_ii}) by varying $\tau$ for the red and blue dots  from the circle class, and the green sample from the triangle class. The circle class and triangle class include 10 samples and 4 samples, respectively.}
\label{fig:soft_assignment_weights}
\end{figure}

In~\cite{ardeshir2022estimating}, the performance of a model across different local neighborhoods in the proxy embedding space is used to estimate disparity across identities/groups. Hence a local neighborhood could be seen as an approximation for a subpopulation/group/identity $g_i$. ~\cite{ardeshir2022estimating} also illustrates that there are different neighborhood sizes that better approximate different group memberships. To capture the same concept, in our formulation
$\tau$ controls the skewness of the exponential function, which influences the size of the local neighborhood. Thus, we fine-tune the hyper-parameter $\tau$ to allow for exploring different neighborhood sizes and therefore different density estimations. Figure~\ref{fig:soft_assignment_weights} shows the impact of $\tau$ on the weight of three different samples. As it can be seen, as $\tau \rightarrow \infty$, the weights converge to $p^\tau_{i} \rightarrow \frac{1}{|\B_{y_i}|}$ which is simply the inverse of per-class frequency.



In the example shown in Figure~\ref{fig:soft_assignment_weights}, the blue and red circles come from the majority {circle} class, and the green triangle comes from the minority class. Often in typical sample weighting schemes, all samples within the same class are weighted uniformly and based on the inverse of their frequency. Thus samples in the minority class are always up-weighted compared to samples in the majority class. However, our sample weighting scheme allows for a more nuanced weighting. Compared with the blue circle, it can be observed that the red circle lies in the denser area, i.e. has more close neighbors. Hence, the $p_i^\tau$ of the red circle is consistently smaller than $p_i^\tau$ of the blue circle for the same $\tau \in (0, \infty)$. Also, they both converge to the inverse frequency $1/10$ when $\tau \rightarrow \infty$. When comparing samples from different classes, the green triangle (which is from the minority class) has more neighbors for smaller $\tau$, and thus could have a smaller weight compared to the blue circle which comes from the majority class. This allows for capturing a more nuanced notion of sample rarity within each class, which goes beyond the typical frequency-based methods. 

Given $p_i^\tau$, the proposed CID method simply minimizing the objective~(\ref{eqn:obj}) such that $p_i^\tau$ is normalized using 
$Z_{y_i}$ to equalize the total contribution of each class. Algorithm~\ref{alg:RBPN-O} describes the practical implementation of the proposed CID method in minimizing the objective~(\ref{eqn:obj}).

\vspace{-0.05in}
\section{Evaluation}
\vspace{-0.05in}
\label{sec:evaluation}
In addition to using standard classification accuracy metrics, we measure the robustness of the trained models using the following metrics. 
\vspace{-0.05in}
\subsection{Area Under DEN Curves (AUD)}
\label{sec:AUDEN}
\vspace{-0.05in}
\cite{ardeshir2022estimating} introduces a metric for estimating structural performance disparity across an embedding space, referred to as \textit{Disparity across Embedding Neighborhoods} (DEN for short). The aforementioned study shows that such a metric is a good estimate of the performance disparity across groups when group information is not available. In fact, our objective function is specifically designed to minimize such disparity, thus we evaluate this metric on the test set to validate our assumption. 
\vspace{-0.05in}
\subsection{Area Under Min-Max Curves (AUMM)}
\label{sec:AUIMMC}
\vspace{-0.05in}
As mentioned earlier, we do not have access to group labels during training, however, to measure if our model is in fact more robust across groups, we use group labels in the test set to validate our hypothesis. In our setup, we mostly focus on robustness/fairness across individuals, and given that the number of individuals in a face dataset could be very large, we define a modification to the widely used Rawlsian min-max metric. In the Rawlsian min max metric~\cite{rawls2001justice}, the ratio of the performance of the model is measured between the most and least accurate groups, i.e. $1-\frac{\min_{g}(e_g)}{\max_{g}(e_g)}|_{g\in G}$. This measure is often very useful when the number of groups is very limited. However, given that in our instance, we are interested in measuring disparity across different people, the number of different individuals in the dataset could be very large. Therefore, using the ratio of performance only on the highest and lowest individual will ignore large portions of the dataset. Thus we modify the Rawlsian min-max formulation to measure the ratio of the bottom-k\% and top-k\% of groups instead. 
\vspace{-0.1in}
\begin{equation}
\text{MM} = \{1 - \frac{\bar{e}_g^k}{ \underline{e}_g^k}\}_{k=1}^{|G|}
\vspace{-0.1in}
\end{equation}
where $k\in [1,\cdots,|G|]$ denotes the index of groups. $\bar{e}_g^k$ and $\underline{e}_g^k$ are the average of top and bottom $k$ group performance, respectively. Sweeping k, results in a curve, which we refer to as the Min Max Curve. We use the area under this curve, AUMM for short,  as a metric for robustness across groups. The lower the AUMM, the more robust/fair the model is.




\begin{algorithm}[t]{\hspace*{-0.5in}}
    \centering
    \caption{CID Optimization ($\tau$)}
    \label{alg:RBPN-O}
    \begin{algorithmic}[1]  
    \STATE Model initialization $\w_1$, proxy embeddings $\{\z_i\}_{i=1}^n$
    \FOR {$t = 1,\ldots, T$}
    \STATE Sample a batch of $B$ samples $\B=\{(\x_i,y_i)\}_{i}^{B}\sim\D$
    \STATE Retrieve the proxy embedding vectors of batch samples, $\{\z_i\}_{i=1}^{B}$.
    \STATE Calculate $p^\tau_{i}$ according to Eqn~(\ref{eqn:p_ii}) for $\forall (\x_i,y_i)\in \B$
    \STATE Calculate $Z_{y_i} = \sum_{j\in \B_{y_i}} p^\tau_{j}$
    \STATE Calculate CID loss: $\sum_{i\in\B}p^\tau_{i}\ell_i(\w_t)/(Z_{y_i})$
    \STATE Update $\w_t$ using stochastic algorithms.
    \ENDFOR
    \STATE {\bf Return} $\w_{T+1}$,
    \end{algorithmic}
\end{algorithm}



\vspace{-0.05in}
\section{Experiments}
\vspace{-0.05in}
We evaluate the proposed approach alongside a few other baselines, on several datasets in terms of overall performance and robustness. In section \ref{sec:datasets} we go over the datasets and tasks used in our evaluation. In section \ref{sec:metrics} we provide details on our evaluation protocol and provide experimental results. Finally, in section \ref{sec:stress_test} we propose and report a stress test to measure robustness to controlled bias. In all experiments, for each face-image in the datasets, we extract its face-recognition embedding vector $\z$ using the face recognition model~\cite{facerecognitionmodel} and use it as its identity proxy. 

\vspace{-0.05in}
\subsection{Datasets and Setup}
\vspace{-0.05in}
\label{sec:datasets}
We selected datasets that contained identity-independent tasks, and also contained information about the identity of the faces in the test set, in order to be able to evaluate the robustness of the trained models across identities at test time. As mentioned earlier, we do not use any identity information during the training phase, and solely rely on off-the-shelf face-recognition embeddings ~\cite{facerecognitionmodel} on the train set. In the following, we provide information on the three datasets used for experiments:\\
\textbf{CelebA}~\cite{liu2015faceattributes} has 200K face images and includes 10117 identities in total. Each image is labeled with 40 attributes/tasks. We pick two identity-independent tasks~\cite{liu2015faceattributes} \{{\it Smiling, Mouse Slightly Open(MSO)}\} and train standard binary classification models to predict those tests. We train ResNet18 model for 20 epochs both on SGD optimizers~\cite{bottou2012stochastic}. The learning rate is tuned in $\{0.003, 0.005, 0.01 \}$ for all the baselines. The hyperparameter $\tau$ in the CID method is tuned in $\{0.1:0.1:0.5 \}$.\\
\textbf{ExpW}~\cite{zhang2015learning} (cleaned version) is a facial expression dataset that includes 85K images with 1002 identities (split into 80\% train, 10\% val, and 10\% test), and 7 labels of facial expressions \{{\it angry, disgust, fear, happy, sad, surprise, neutral}\}. 
To balance the size of the data scale and model capacity, we combine {\it sad-surprise-fear-neutral } as the new {\it ssfn} attribute to have enough positive samples ($\sim$ 19\%) to learn a valid CNN model. Then we predict \{{\it angry, disgust, happy, ssfn}\} expressions, respectively. Following the experimental setup in ~\cite{wang2019symmetric}, we adopt the SGD optimizer to optimize all a 4-layers CNN model. The structure of the model is provided in the appendix. We train 40 epochs using SGD optimizer. The learning rate is tuned $\{0.1, 0.05, 0.01\}$ and decayed at the 20th epoch by a factor of 100. $\tau$ is tuned in $\{0.1:0.1:0.5 \}$.
\\
\textbf{PugFig}~\cite{kumar2009attribute}\footnote{54K images of downloading URLs have been provided for the original dataset. But, most of them failed.} including 9K images with 111 identities (split into 80\% train, 10\% val, and 10\% test). Each image is tagged with the binary labels of lighting position, \{{\it frontal, non-frontal}\} and facial expression, \{{\it neutral, non-neutral}\}.
We train models to predict each task separately. Due to the limited data scale, we train the PubFig on the pre-trained ResNet18 model and fine-tune the fully connected layer for 60 epochs using SGD optimizer. The batch size is 16 and the weight-decay parameter is 5e-4. The learning rate is tuned in $\{0.005:0.001:0.01 \}$ and decayed at the 30th epoch by a factor of 10. $\tau$ is tuned in $\{0.1:0.1:0.5 \}$.\\

\vspace{-0.1in}
\subsection{Baselines}
\label{sec:baselines}
\vspace{-0.05in}
We compare proposed approach (CID) with the following effective baselines on fairness under awareness setup:

\textbf{IFW} (inverse frequency weighting)~\cite{huang2016learning,  wang2017learning}: This is the typical sample weighting used to enforce class balance. Given $N_p, N_n$ number of positive and negative samples, the weights for the positive class and negative class samples are set proportional to $1/N_p$ and $1/N_n$, respectively. 

\textbf{DRO} (Distributionally Robust Optimization)\cite{li2021tilted,qi2020attentional}: We implement the ABSGD~\cite{qi2020attentional} stochastic optimization method for optimizing $\max_\p\sum p_i\ell_i(\w)-\lambda\sum{p_i}\log np_i$ The hyperparameter $\lambda$ for DRO is tuned in $\{0.1, 0.5, 1, 2, 5\}$

\textbf{IRM} (Invariant Risk Minimization): we optimize $\ell(\w) + \lambda \|\nabla_{\upsilon|\upsilon=1.0} \ell(\upsilon\cdot \w)\|^2$ objective using the optimization framework of~\cite{arjovsky2019invariant}. $\lambda$ for IRM is tuned in $\{0.1:0.1:1\}$.

\textbf{ARL} (Adversarial Reweighted Learning)\cite{lahoti2020fairness}:
 ARL optimizes $\min\limits_{\w}\max\limits_{\phi}p_i^{\phi} \ell_i(\w)$, where $p_i^\phi = 1/n +f_i^{\phi}/\sum_i f_i^{\phi}$, where $\phi$ is a linear adversary model with output score $f_i^\phi$\footnote{The details of implementation are provided in the supplimentary.}.



\vspace{-0.05in}
\subsection{Measuring Robustness} 
\vspace{-0.05in}
\label{sec:metrics}
We evaluate the performance of the baselines mentioned in Section \ref{sec:baselines}, in terms of overall classification accuracy (Acc), average, and standard-deviation of per-identity accuracy (Id Acc and $\delta_{\text{Id}}$). This is evaluated by measuring a model's prediction accuracy for each identity (person) in the test set, and reporting mean and standard deviation. Ideally a high-performing and identity-robust model should maintain high Id Acc, while having low $\delta_{\text{Id}}$. A low $\delta_{\text{Id}}$ is one of the metrics implying that the performance of the model is robust across identities and thus more fair. In addition, we report the accuracy on the least accurate 10\% of identities in the test set. Plus, we evaluate the model in terms of area under DEN curve (referred to as AUD for short), proposed in \cite{ardeshir2022estimating}, and explained in section \ref{sec:AUDEN}, which measures the disparity of performance across different neighborhoods of the embedding space of the face-recognition embedding. The lower the AUD, the more robust the model is. Also, as described in section \ref{sec:AUIMMC}, we use the area under the min-max curve (AUMM for short) as another robustness metric. Given that this metric measures the disparity between the accuracy of the top-k and bottom-k identities, the lower the AUMM, the more identity-robust a model is.

Table~\ref{tab:ID_CelebA}, \ref{tab:expW}, and \ref{tab:pubfig} report  experimental results on CelebA, ExpW, and PubFig datasets respectively, and averaged over 5 independent runs. It can be observed that CID outperforms all the baselines in terms of every model robustness metric, namely, (lowest) 10\% Id Acc, $\delta_{\text{Id}}$, AUMM, and AUD. In addition, the overall accuracy is also either higher than baseline, or very competitive. In other words, CID does maintain high accuracy while attaining robustness.

\begin{figure}[htbp]
    \centering
 \includegraphics[width = 0.45\linewidth]{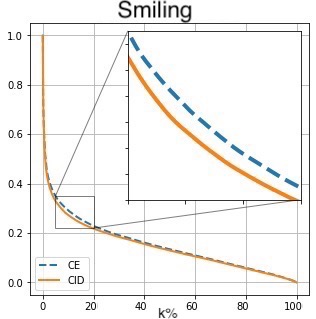}
\includegraphics[width = 0.45\linewidth]{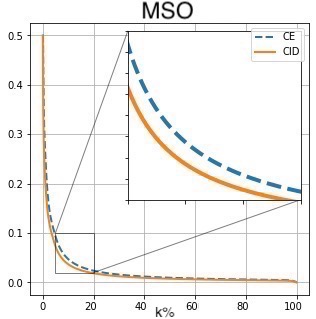}
    \caption{The Min-Max curves for the {\it Smiling} and {\it Mouth Slightly Open (MSO)} tasks. It can be observed that CID consistently yields a lower curve, resulting in a smaller area under the MM curve, and thus, less disparity across top and bottom k\% identities.}
\label{fig:IMMC_Default_Data}
\end{figure}

\begin{table}[h!]
\caption{Attributes Prediction Experimental Results on CelebA. {\bf Bold} and \underline{underline} are  best and second-best results of each metric.}
\label{tab:ID_CelebA}
\centering
 \resizebox{\linewidth}{!}{ 
\begin{tabular}{l|c|c|c|c|c|c}
\toprule
  {\bf Smiling}     & Acc & Id Acc & 10\% Id Acc & $\delta_{\text{Id}}$ & AUMM & AUD \\ \hline
CE      &  \textbf{92.73}  &    \underline{92.02}  &    71.02           &       0.0962       &  \underline{0.1544} & 0.1156  \\
IFW     &  92.69   &    \textbf{92.99}    &        71.03    &       0.0968         &    0.1550 & \underline{0.1141}  \\
DRO     &   92.55  &   91.90     &      71.06    &  0.0965            &   0.1546 & 0.1146 \\
IRM     & 92.71    &  92.11      &    \underline{ 71.74  }    &        0.0939      & 0.1522   &  0.1143 \\ 
ARL     & 92.71   & 92.07  &    70.94  &        0.0975   & 0.1558 &  0.1151 \\ \hline
CID     &\underline{92.72}     &     \underline{92.15}     &   \textbf{71.94}           &           \textbf{0.0926}    &    \textbf{0.1506} & \textbf{0.1126}\\
\bottomrule
\end{tabular}}
\resizebox{\linewidth}{!}{ 
\begin{tabular}{l|c|c|c|c|c|c}
\toprule
  {\bf MSO }    & Acc & Id Acc & 10\% Id Acc & $\delta_{\text{Id}}$ & AUMM & AUD \\ \hline
CE      &   94.09  &  93.71      &       74.57      &    0.0838          &   0.1322 &0.1036 \\
IFW     &  \underline{94.09}   & \underline{93.73}       &  75.06          &   \underline{0.0832}         &  \underline{0.1305} & \underline{0.1019}  \\
DRO     &   94.04  & 93.69       &   75.07          &      0.0840        &   0.1341 &0.1038 \\
IRM     &    94.04 &    92.68    &        \underline{75.10}     &       0.0837       &  0.1325 &  0.1021 \\
ARL     & 93.97   &  93.54  &   74.04   &        0.0871    & 0.1376  &  0.1033 \\ 
\hline
CID &  {\bf 94.13}   &  \textbf{93.79}      &     \textbf{75.36}        &    \textbf{0.0824}         & \textbf{0.1299} & \textbf{0.1005 } \\
\bottomrule
\end{tabular}
}
\end{table}

\begin{table}[h!]
\caption{Expression Prediction Experimental Results on ExpW.  }
\label{tab:expW}
\centering
\resizebox{\linewidth}{!}{ 
\begin{tabular}{l|c|c|c|c|c|c}
\toprule
  {\bf Angry}      & Acc & Id Acc & 10\% Id Acc & $\delta_{\text{Id}}$ & AUMM & AUD \\ \hline
CE      &   \underline{63.25} &    \underline{ 63.26}  &        \underline{ 33.46}    &     0.1706         &   0.3603 & 0.2270 \\
IFW     &  60.83 &  60.84   &   32.29          &         \underline{0.1691}    &   0.3643 & 0.2309\\
DRO     & 63.18  &    63.19   &     33.26    &        0.1693      & 0.3622 & \underline{0.2205}  \\
IRM     & 63.17     &     63.18   &        33.20     &      0.1712        &  0.3615  & 0.2232 \\
ARL     & 63.44    &    63.45  &       33.43     &      0.1693       &  \underline{0.3584} & 0.2214 \\\hline
CID     &  \textbf{ 63.53} & \textbf{63.53}    &    \textbf{33.53 }          &       \textbf{0.1672}        &  \textbf{0.3543}& \textbf{0.2188}\\
\bottomrule
\end{tabular}}
\resizebox{\linewidth}{!}{ 
\begin{tabular}{l|c|c|c|c|c|c}
\toprule
  {\bf Disgust}     & Acc & Id Acc & 10\% Id Acc & $\delta_{\text{Id}}$ & AUMM & AUD \\ \hline
CE      & 75.23    &   75.23     &            \underline{ 46.19} &             0.1517 & \underline{0.2862}  & 0.1875    \\
IFW     &  74.93  & 74.94       &        44.99     &            \underline{0.1514} &  0.2938 & \underline{0.1758}\\
DRO     &  \underline{75.26}   & \underline{ 75.27}      &     46.09        &   0.1520           &    0.2865 & 0.1867 \\
IRM     &  75.25   &    75.27   &    46.00         &        0.1518      &   0.2863 &  0.1921\\
ARL     & 75.13  &   75.14   &    45.80        &        0.1531     &   0.2893 &  0.1911\\\hline
CID     &  \textbf{75.27}   &  \textbf{75.28}     &           \textbf{46.22}  &          \textbf{0.1503}   &  \textbf{0.2837}  & \textbf{0.1749}  \\
\bottomrule
\end{tabular}}

\resizebox{\linewidth}{!}{ 
\begin{tabular}{l|c|c|c|c|c|c}
\toprule
{\bf Happy}    & Acc & Id Acc & 10\% Id Acc & $\delta_{\text{Id}}$ & AUMM & AUD\\ \hline
CE      &  86.94  &   86.95    &  59.28      &    0.1292         &  0.2159 & 0.1429
 \\
IFW     &   86.86   &    86.87       &     59.24       &     0.1290           & 0.2160  & 0.1326 \\
DRO     &   86.95  &    86.96   &  59.34           &         \underline{0.1286}    &  0.2151 & \underline{0.1311} \\
IRM     &  \underline{86.98}  &  \underline{87.00}      &  \underline{59.73}           &         0.1289     &   \underline{0.2148} & 0.1386\\ 
ARL     &  86.73  &  86.95      &  0.5957          &         0.1290    &   0.2166 & 0.1391\\ \hline
CID     &  \textbf{87.00}   &   \textbf{87.01}     &           \textbf{59.99}  &      \textbf{0.1281}      &  \textbf{0.2140}  & \textbf{0.1251} \\
\bottomrule
\end{tabular}}
\resizebox{\linewidth}{!}{ 
\begin{tabular}{l|c|c|c|c|c|c}
\toprule
  {\bf SSFN}   & Acc & Id Acc & 10\% Id Acc & $\delta_{\text{Id}}$ & AUMM & AUD \\ \hline
CE      &  \underline{85.16}  &  \underline{85.17}    &      58.85   &        0.1273    &  \underline{ 0.2187} & 0.1583\\
IFW     & 79.98  &    79.99     &    53.13       &       0.1434         & 0.2588 &   0.1695 \\
DRO     &   85.19  &     85.20   &      58.95    &    0.1278          & 0.2200   & \underline{0.1583}\\
IRM     &   \textbf{85.20}  &    \textbf{ 85.21}   &     \underline{59.08}    &    \underline{0.1271}          & 0.2197  & 0.1632 \\
ARL     &   85.07  &    84.51   &     57.52    &    0.1342          & 0.2259  & 0.1655 \\\hline
CID     &  85.14  &    85.15   &     \textbf{59.60}        &    \textbf{0.1260}          &   \textbf{0.2163}  & \textbf{0.1576}\\
\bottomrule
\end{tabular}}
\end{table}

\begin{table}[h!]
\caption{Experimental Results on PubFig. }
\label{tab:pubfig}
\centering
 \resizebox{\linewidth}{!}{ 
\begin{tabular}{l|c|c|c|c|c|c}
\toprule
  {\bf Frontal}      & Acc & Id Acc & 10\% Id Acc & $\delta_{\text{Id}}$ & AUMM & AUD \\ \hline
CE      &   74.83  &    74.93  &  36.11           &       \underline{0.1913}      &  0.3487  & \underline{0.1639} \\
IFW     & 71.68    & 71.77  &         34.72    &         0.1969      &   0.3582   & 0.1837 \\
DRO     & 74.60    &   74.70     &      37.15       &   0.1930           &   0.3491 & 0.1660\\
IRM     & \underline{75.13}    &              \underline{75.24}  &        \underline{37.85} & 0.1974      &  \underline{0.3465}  & 0.1657 \\
ARL     & 73.56    &   74.77   &      37.85       &   0.1931           &   0.3479 & 0.1655\\ \hline
CID     & \textbf{75.10}     &  \textbf{75.20}     &  \textbf{ 38.19}         &           \textbf{ 0.1906 }  &     \textbf{ 0.3449} & \textbf{0.1606}\\
\bottomrule
\end{tabular}}
\resizebox{\linewidth}{!}{ 
\begin{tabular}{l|c|c|c|c|c|c}
\toprule
  {\bf Neutral}   & Acc & Id Acc & 10\% Id Acc & $\delta_{\text{Id}}$ & AUMM & AUD \\ \hline
CE      &  56.48 &  56.55    &       26.74     &    0.1798         &   \underline{0.3985} & 0.2645 \\
IFW     &  \underline{56.61}  & \textbf{56.69}       & 26.74      &     \underline{0.1773}       &  0.4050   & 0.2593 \\
DRO     &  56.24  & 56.32      &   26.74          &      0.1782      &   0.4034 & 0.2588 \\
IRM     &  56.48   &    56.55   &     \underline{27.08}      &    0.1817          &    0.4009  & \underline{0.2582}\\ 
ARL    &  \textbf{56.91}  &    55.85  &   26.04      &   0.1772       &    0.3995  & 0.2575\\
\hline
CID     & 56.51  & \underline{56.59}    &    \textbf{  27.43}         &          \textbf{0.1677}       & \textbf{0.3864} &  \textbf{0.2544} \\
\bottomrule
\end{tabular}
}
\end{table}


\vspace{-0.1in}
\subsection{Stress-testing with Controlled Bias}
\begin{figure*}[t!]
    \centering
\includegraphics[width = 0.21\linewidth]{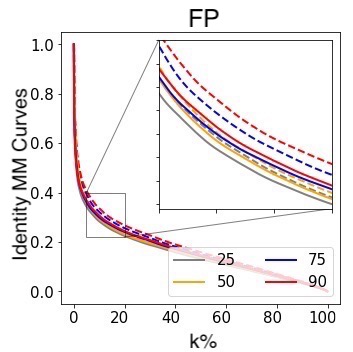}
\includegraphics[width = 0.18\linewidth]{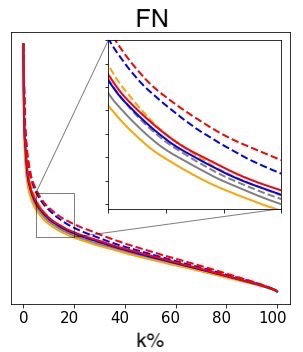}
\includegraphics[width = 0.18\linewidth]{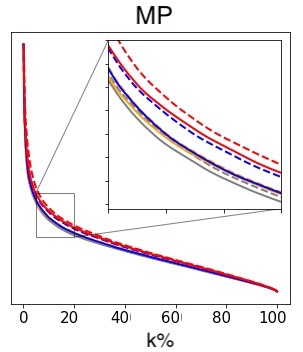}
\includegraphics[width = 0.18\linewidth]{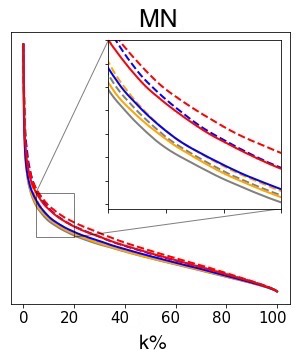}
\includegraphics[width = 0.18\linewidth]{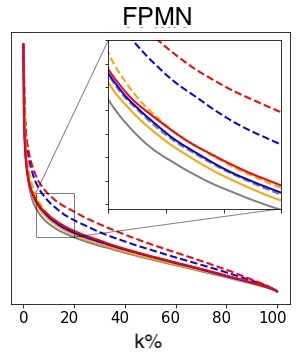}

\includegraphics[width = 0.22\linewidth]{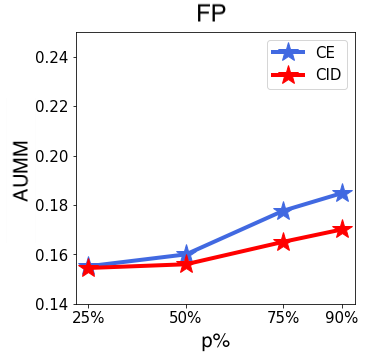}
\includegraphics[width = 0.18\linewidth]{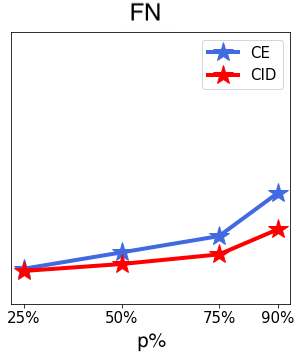}
\includegraphics[width = 0.18\linewidth]{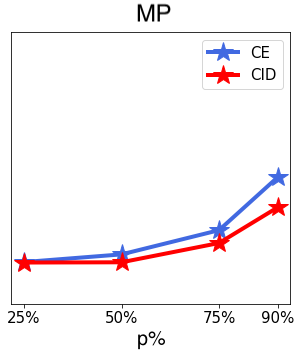}
\includegraphics[width = 0.18\linewidth]{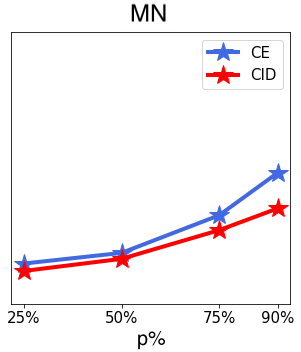}
\includegraphics[width = 0.18\linewidth]{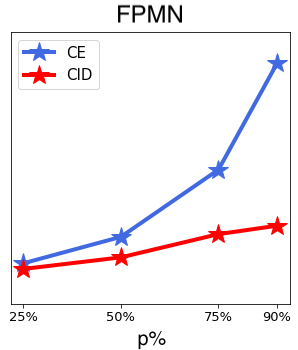}

\includegraphics[width = 0.225\linewidth]{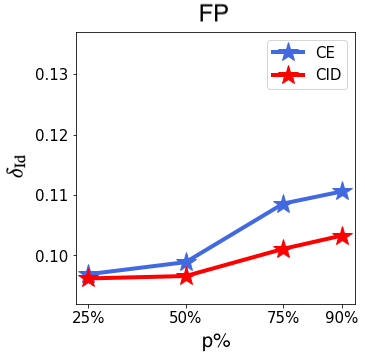}
\includegraphics[width = 0.18\linewidth]{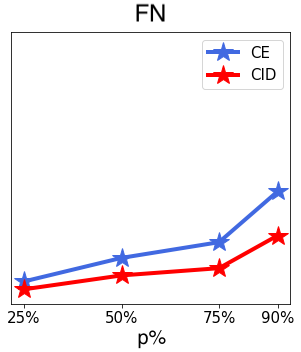}
\includegraphics[width = 0.18\linewidth]{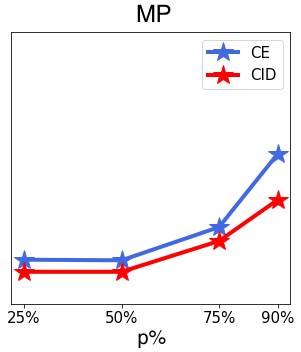}
\includegraphics[width = 0.18\linewidth]{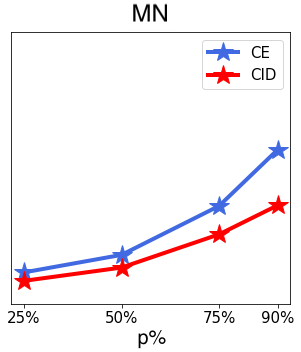}
\includegraphics[width = 0.18\linewidth]{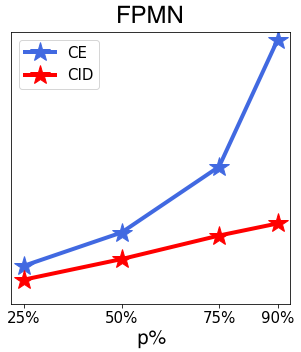} 

\includegraphics[width = 0.22\linewidth]{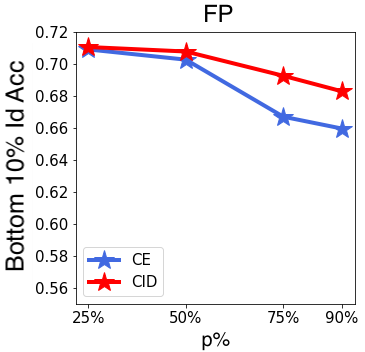}
\includegraphics[width = 0.18\linewidth]{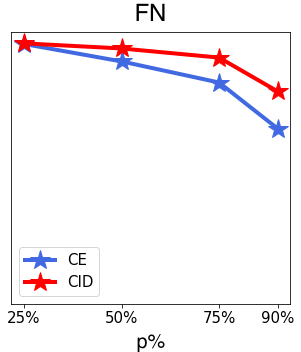}
\includegraphics[width = 0.18\linewidth]{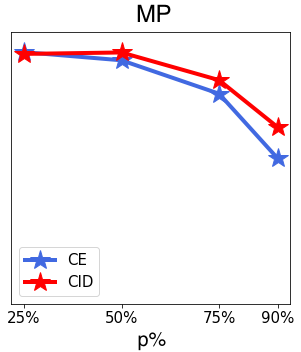}
\includegraphics[width = 0.18\linewidth]{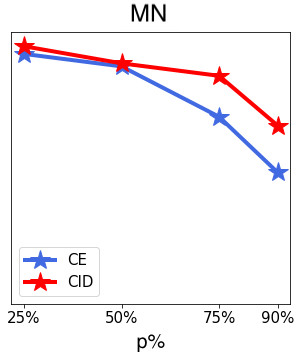}
\includegraphics[width = 0.18\linewidth]{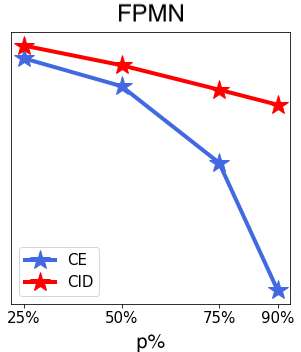}
\caption{Results of different stress tests on the CelebA dataset for the smiling task label. The identity MM curves, area under identity min-max curves (AUMM), standard deviation of Id accuracy $\delta_{\text{Id}}$, and the bottom 10\% percentage Id accuracy are measured for different setups and under different levels of induced bias. First row: For the MM figure, the x-axis shows k for which the disparity between top and bottom k identities is evaluated. the In each figure, x-axis specifies the amount (percentage) of the training data of the (group, task-label) that is excluded during training. As it can be observed in all instances. As it can be observed, CID maintains its original metrics significantly better than CE in the presence of distribution shift.}
\label{fig:OOD_results}
\end{figure*}

\vspace{-0.05in}
\label{sec:stress_test}
In this section, we aim to measure how sensitive to dataset bias a model is by stress testing the model. To do so, we construct different versions of the train/validation sets of CelebA, by manipulating the dataset and adding controlled artificial identity-to-task bias. More specifically, given a task (such as smiling), we construct a biased train set by excluding $p$\% of data points belonging to a (task-label, subpopulation). As an example, in one of the variations, we exclude 50\% of male-smiling images. This would artificially create a biased dataset that is prone to correlating male faces to the label non-smiling. If we train a classifier on such a dataset, a model's performance on the standard (non-manipulated) test set can be very non-robust, as the train and test set do not follow the same distribution. In all the setups we solely manipulate the bias in the train set and keep the test set unchanged. We try this experiment for different values of $p \in \{25\%, 50\%, 75\%, 90\%\}$, and for different (group, task-label) combinations. We refer to each setting by specifying which group (M:Male vs. F:Female), and which task label (P: positive, N: negative) has been manipulated (excluded by p\%) from the training and validation set (while the test set is unchanged). As an example, on the task "Smiling", FP 50\% means that half of the Female Positives (smiling female faces) were excluded during training, therefore creating a bias in the dataset. Combinations of these setups could also be generated to further exaggerate the bias (such as FPMN: removing positive/smiling female images and negative/non-smiling male images). Figure~\ref{fig:OOD_results} shows the results of our stress test on the task "Smiling". Due to space limitations, we only report the robustness measures, namely, MMC, $\delta_{\text{Id}}$, AUMM, and bottom 10\% Id Acc experimental results between CE (cross-entropy) and CID on all different setups and levels of induced bias. We also provide more stress test results, including other metrics, analysis on the other CelebA task, and other variations of the manipulation setups in the supplementary. All other variations of tasks, metrics and setups follow the same trend.


\begin{table}[htbp]
\caption{Stress testing the models on the CelebA dataset, by eliminating 90\% of a subpopulation in the training/validation set.}
\label{tab:OOD_CelebA}
\centering
\resizebox{\linewidth}{!}{ 
\begin{tabular}{l|c|c|c|c|c|c}
\toprule
  {\bf FP}      & Acc & Id Acc & 10\% Id Acc & $\delta_{\text{Id}}$ & AUMM  & AUD \\ \hline
CE      &    90.33 &      89.77  & 65.91            &          0.1095    &    0.1821 &0.1354 \\
IFW     &  91.13   &    \underline{90.49}    &\underline{67.28}             &       \underline{0.1053}       &  \underline{0.1737} &  \underline{0.1259}\\
DRO     &  90.23   &    89.84    &  66.20           &    0.1098          &    0.1834 & 0.1367\\
IRM     &  \underline{91.18}  &     90.11   &       66.54      &         0.1088     &  0.1789   & 0.1343 \\
ARL     &   90.40  & 89.46   &        65.55    &      0.1127      &  0.1863  & 0.1323  \\ \hline
CID     &  \textbf{91.31}   &  \textbf{90.67}       &      \textbf{67.73}       &      \textbf{0.1042}       &  \textbf{0.1717} & \textbf{0.1233}\\
\bottomrule
\end{tabular}}
\resizebox{\linewidth}{!}{ 
\begin{tabular}{l|c|c|c|c|c|c}
\toprule
 {\bf FN}     & Acc & Id Acc & 10\% Id Acc & $\delta_{\text{Id}}$ & AUMM & AUD \\ \hline
CE      &  89.42   &   89.51     &   66.52          &  0.1077            &   0.1821 & 0.1476 \\
IFW     & \underline{90.47}   &   \underline{ 90.29}    &         \underline{67.45}    &  \underline{0.1074}            & \underline{0.1761} & \underline{0.1281}   \\
DRO     &   89.43  &    89.51    &    66.53         &  0.1077            & 0.1817 & 0.1501   \\
IRM     &   90.18  &     89.91   &       66.84      &         0.1075     &  0.1799  & 0.1438 \\ 
ARL     &   89.64  &  89.66    &        66.78    &     0.1075      &  0.1807  & 0.1456  \\ \hline
CID     &  \textbf{90.89}   &    \textbf{90.58}    &       \textbf{68.12}    &  \textbf{0.1057}        & \textbf{0.1727} & \textbf{0.1241}   \\
\bottomrule
\end{tabular}}

\resizebox{\linewidth}{!}{ 
\begin{tabular}{l|c|c|c|c|c|c}
\toprule
 {\bf MP  }  & Acc & Id Acc & 10\% Id Acc & $\delta_{\text{Id}}$ & AUMM & AUD\\ \hline
CE      &  \underline{91.51}   &  90.03      &      64.50       &          0.1162    &  0.1885 &  0.1349  \\
IFW     & 91.34    & \underline{90.35}       &      \underline{65.69}      &         \underline{0.1127}    & \underline{ 0.1820} & \underline{0.1309} \\
DRO     & 91.12    &    90.13    &         64.70    &  0.1172         &    0.1855 & 0.1381   \\
IRM     &  91.22   &   90.23     &       65.01      &          0.1168    & 0.1823  & 0.1361    \\ 
ARL     &   91.27  &  90.17    &      64.49    &      0.1169    &   0.1890  & 0.1371  \\ \hline
CID     &    \textbf{91.58} &  \textbf{90.62}      &        \textbf{66.68}     &         \textbf{0.1091}     &   \textbf{0.1774} & \textbf{0.1221} \\
\bottomrule
\end{tabular}}

\resizebox{\linewidth}{!}{ 
\begin{tabular}{l|c|c|c|c|c|c}
\toprule
 {\bf MN}     & Acc & Id Acc & 10\% Id Acc & $\delta_{\text{Id}}$ & AUMM & AUD\\ \hline
CE      & 90.29 &  89.53      &    63.72         &          0.1151    & 0.1922  & 0.1433  \\
IFW     &   90.07  &      \underline{90.24}  &     \underline{65.68}       &   \underline{0.1093}          &  \underline{0.1816} & \underline{0.1372}  \\
DRO     &  90.03   &     89.60   &    63.91         &      0.1146        &  0.1915  & 0.1490  \\
IRM     &  90.06   &  89.70       &      64.11       &          0.1123    &  0.1845    & 0.1455\\ 
ARL     &    \underline{90.33}   &89.51   &        63.54    &      0.1162      &  0.1956  & -  \\ \hline
CID     & \textbf{91.20}    &   \textbf{90.43}     &   \textbf{66.16}          &           \textbf{0.1080}   &  \textbf{0.1783}   & \textbf{0.1241}\\
\bottomrule
\end{tabular}}
\resizebox{\linewidth}{!}{ 
\begin{tabular}{l|c|c|c|c|c|c}
\toprule
  {\bf FPMN }    & Acc & Id Acc & 10\% Id Acc & $\delta_{\text{Id}}$ & AUMM & AUD\\ \hline
CE      & 87.25  &   86.48    &   57.28        &      0.1295     &  0.2229  & \underline{0.1584} \\
IFW     &   87.23   &    86.41       &       57.66        &     0.1322           & 0.2279   & 0.1655 \\
DRO     &  \underline{87.37}   &     \underline{86.62}   &        
\underline{59.14} &      \underline{0.1280}        &   \underline{0.2208} & 0.1625 \\
IRM     &   87.36  &  86.63      &        59.13     &        0.1283      &  0.2210  & 0.1614  \\ 
ARL     &   87.32  &  86.55    &        58.33    &      0.1303      &  0.2247  & 0.1631  \\ \hline
CID     & \textbf{87.87}    &   \textbf{87.09}     & \textbf{60.24}            &   \textbf{ 0.1264}          & \textbf{0.2170}  &\textbf{0.1553}\\
\bottomrule
\end{tabular}}
\end{table}

\begin{table}[h!]
\centering
\caption{Accuracy of the model on the least accurate subpopulation in the CelebA dataset \{{\it task-label$\times$Male/Female}\} on the stress testing CelebA dataset. It can be observed that CID achieves the highest worst-case accuracy, and therefore the most robust among the evaluated baselines.}

\label{tbl:worst_group_domain_shifts}
\resizebox{0.7\linewidth}{!}{ 
\begin{tabular}{c|c|c|c|c|c}
\toprule
    & FP    & FN    & MP    & MN    & FPMN  \\ \hline
CE  & 83.32 & 76.96 & 72.87 & 78.09 & \underline{76.24} \\
IFW &\underline{87.87} &\underline{80.24} &\underline{ 75.46} & \underline{82.46} & 73.34 \\
DRO & 83.16 & 76.98 & 73.05 & 78.40 & 76.19 \\
IRM & 83.77 & 77.04 & 73.51 & 78.33 & 75.93 \\
ARL & 83.09 & 76.47 & 74.37 &78.32 & 75.62 \\\hline
CID & \textbf{88.26} & \textbf{82.95} & \textbf{77.51} & \textbf{83.88} & \textbf{76.47} \\ 
\bottomrule
\end{tabular}}
\vspace{-0.1in}
\end{table}

We observe that training models on these biased versions would cause their test performance to reduce (degrade) as the amount of bias increases. Similarly, disparity metrics increase (degrade) with more bias, as expected. This shows that our stress-testing framework in fact does introduce controlled bias to the trained model. We make the following observations from these experiments: 1)
For each \{{\it task-label $\times$ Male/Female}\} combination, the higher the amount of manipulation, the higher the MMC  and AUMM values, which implies that the proposed metrics (MMC and AUMM) do capture the amount of bias in a model. 2) The proposed CID method has smaller MMC, AUMM, and $\delta_{\text{Id}}$ values, in addition to higher bottom 10\% Id accuracy compared to CE. The gap between the models steadily increases as the amount of induced bias in the dataset increases, which verifies the advantages of CID on handling distribution shifting over Empirical Risk Minimization.
To narrow down the scope, we report experimental results that compare CID with more baseline methods on the most biased version of each stress-test setup (i.e, MP 90\%, MN 90\%, FP 90\%, FN 90\%, and FPMN 90\% in Table~\ref{tab:OOD_CelebA}). Please note, given that we need access to group labels such as Male/Female and identity labels, we were only able to conduct this specific experiment on the CelebA dataset. To conclude this stress-test, all robustness metrics are significantly more desirable for CID compared to the baselines, alluding to the fact that our CID weighting scheme successfully mitigates bias, and maintains high accuracy in presence of it. To make our experiments complement, we report the worst group accuracy that is widely used in the baseline comparison in terms of \{{\it task-label $\times$ Male/Female}\} in Table~\ref{tbl:worst_group_domain_shifts}.

\vspace{-0.1in}
\section{Conclusion}
\label{sec:conclusion}
\vspace{-0.1in}
We propose a framework to effectively use off-the-shelf face-recognition model embeddings to improve the robustness/fairness of identity-independent face models. Our experiments show that our simple sample-weighting approach helps face models to maintain high accuracy while gaining significant robustness to distribution shifts and different levels of bias, and often maintaining a more uniform performance across different identities (and groups) of faces.

\newpage

{\small
\bibliographystyle{ieee_fullname}
\bibliography{egbib}
}

\newpage
$ \ \ \ $
\newpage
\section{Appendix}
Due to space limitations, we omitted including some additional experimental results pertaining to stress tests, which we provide here in Section \ref{sec:stress_test_cont}. We include an additional ablation study on the effect of the hyper-parameter $\tau$, on our robustness metric AUMM, $\delta_{\text{Id}}$, and accuracy in Section \ref{sec:ablation_tau}. In Section~\ref{sec:Aba_RMM}, we report the Rawlsian min-max (RMM) results on identity-related gender and age groups for different methods. 
Finally, we describe the detailed implementation of ARL, and provide connections and comparisons between CID and baselines in Section~\ref{sec:cc_baselines}.

\subsection{More Results on the stress-tests}
\label{sec:stress_test_cont}
\begin{figure*}[htbp]
    \centering
\includegraphics[width = 0.19\linewidth]{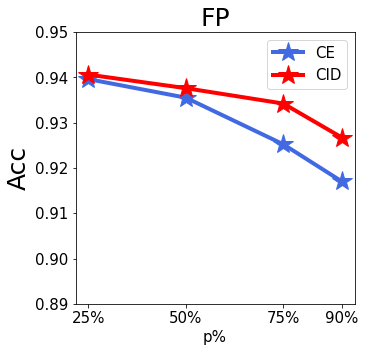}
\includegraphics[width = 0.19\linewidth]{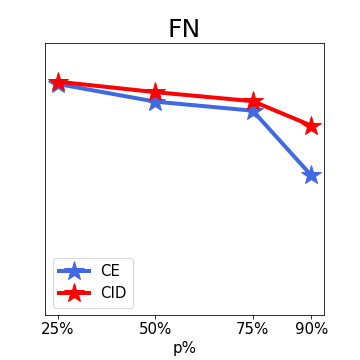}
\includegraphics[width = 0.19\linewidth]{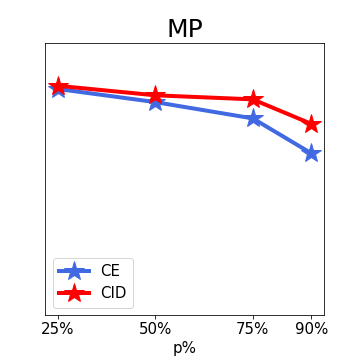}
\includegraphics[width = 0.19\linewidth]{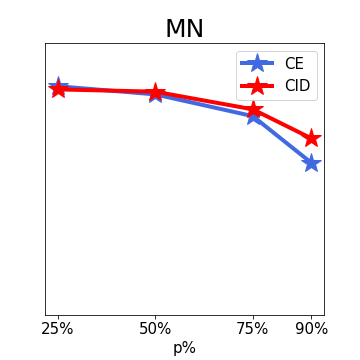}
\includegraphics[width = 0.19\linewidth]{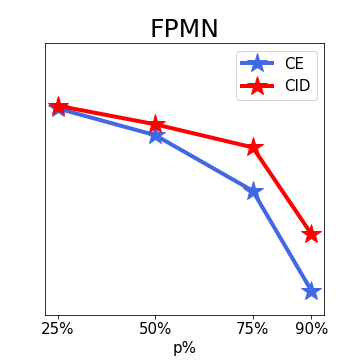}

\includegraphics[width = 0.19\linewidth]{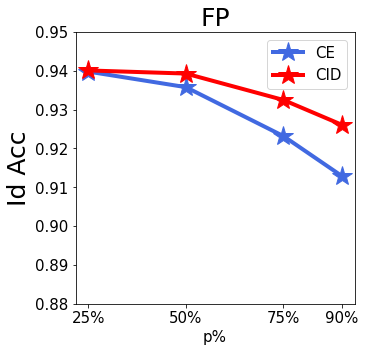}
\includegraphics[width = 0.19\linewidth]{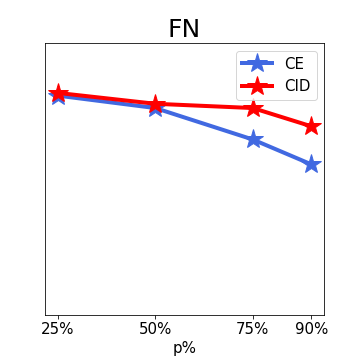}
\includegraphics[width = 0.19\linewidth]{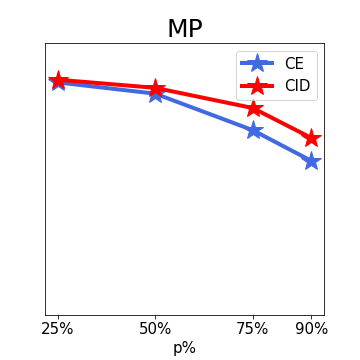}
\includegraphics[width = 0.19\linewidth]{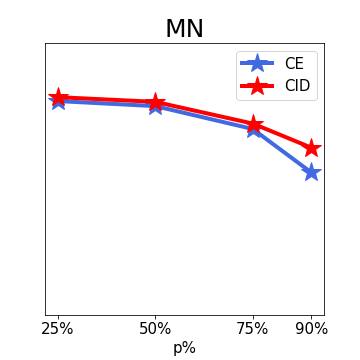}
\includegraphics[width = 0.19\linewidth]{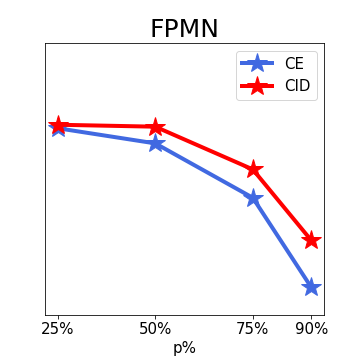}

\includegraphics[width = 0.19\linewidth]{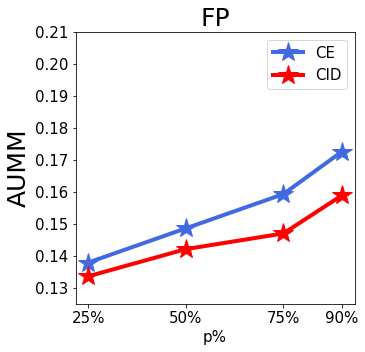}
\includegraphics[width = 0.19\linewidth]{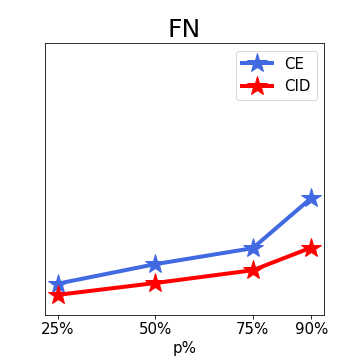}
\includegraphics[width = 0.19\linewidth]{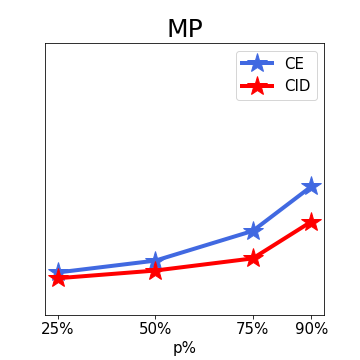}
\includegraphics[width = 0.19\linewidth]{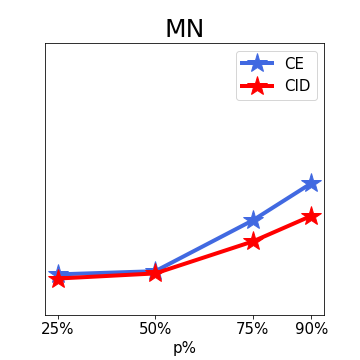}
\includegraphics[width = 0.19\linewidth]{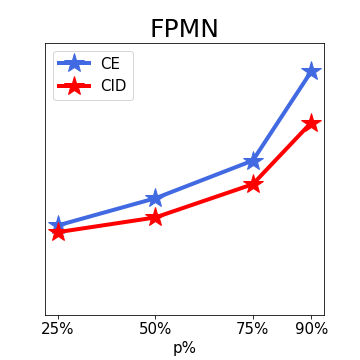}

\includegraphics[width = 0.19\linewidth]{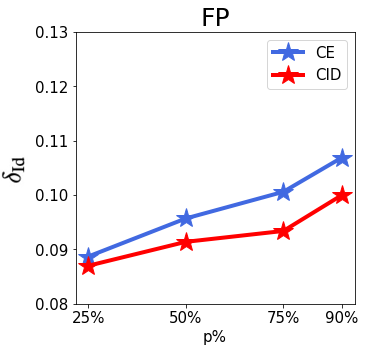}
\includegraphics[width = 0.19\linewidth]{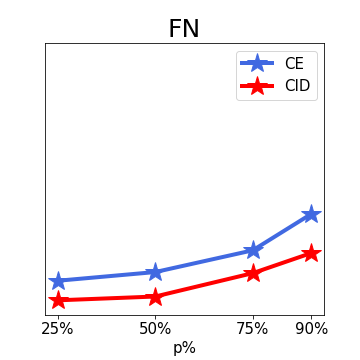}
\includegraphics[width = 0.19\linewidth]{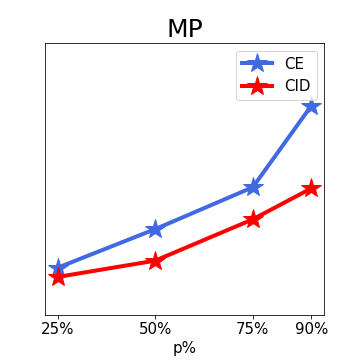}
\includegraphics[width = 0.19\linewidth]{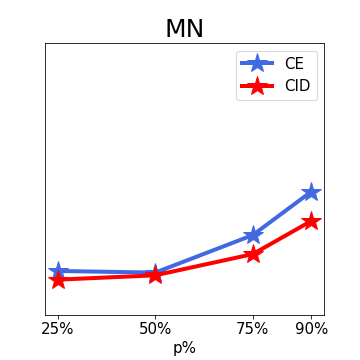}
\includegraphics[width = 0.19\linewidth]{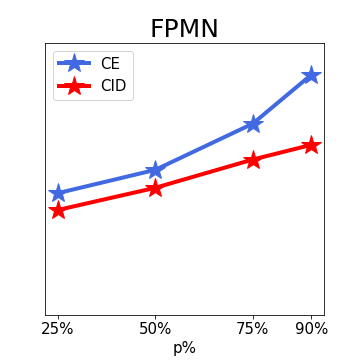}

\includegraphics[width = 0.19\linewidth]{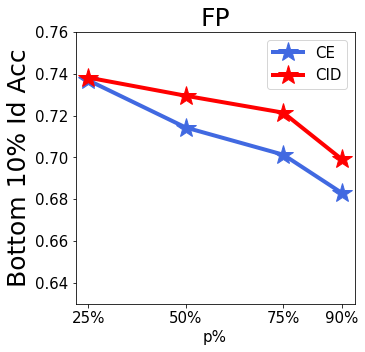}
\includegraphics[width = 0.19\linewidth]{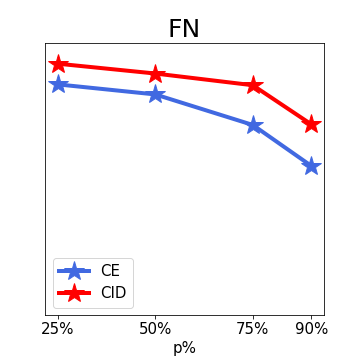}
\includegraphics[width = 0.19\linewidth]{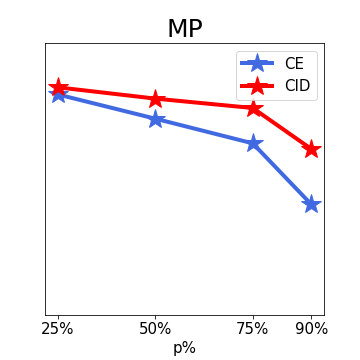}
\includegraphics[width = 0.19\linewidth]{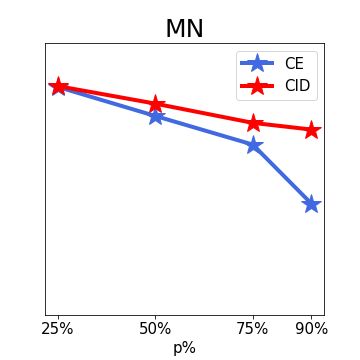}
\includegraphics[width = 0.19\linewidth]{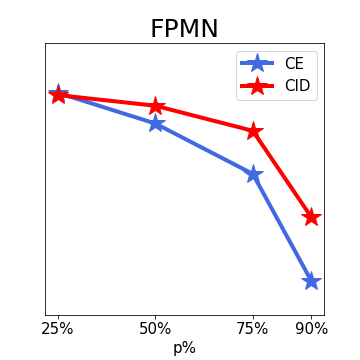}

\caption{Results of different stress tests on the CelebA dataset for the mouth slightly open (MSO) task. We observe trends similar to the ones in the smiling task reported in Figure \ref{fig:OOD_results}. CID seems consistently more robust, and less sensitive to bias compared to CE.}
\label{fig:MSO_stress_test}
\end{figure*}

\begin{figure*}[t!]
    \centering
\includegraphics[width = 0.19\linewidth]{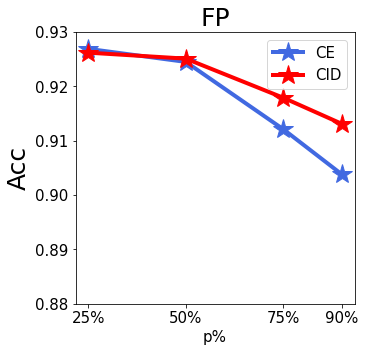}
\includegraphics[width = 0.19\linewidth]{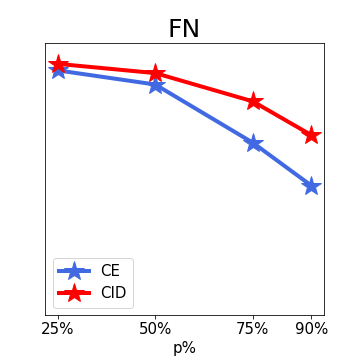}
\includegraphics[width = 0.19\linewidth]{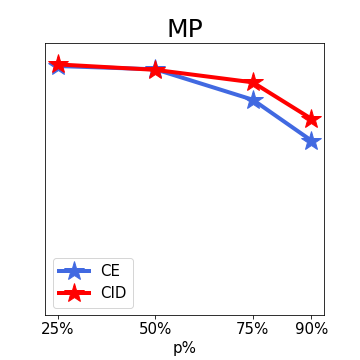}
\includegraphics[width = 0.19\linewidth]{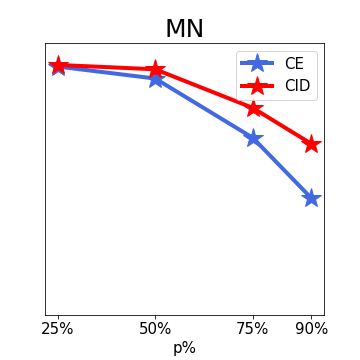}
\includegraphics[width = 0.19\linewidth]{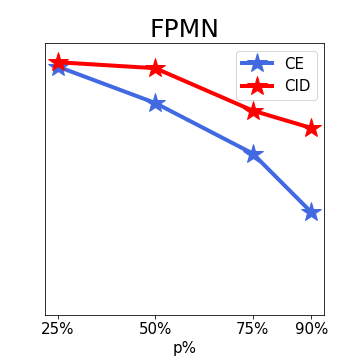}

\includegraphics[width = 0.19\linewidth]{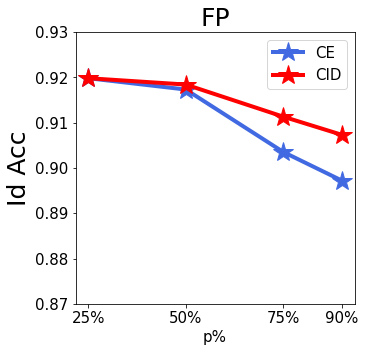}
\includegraphics[width = 0.19\linewidth]{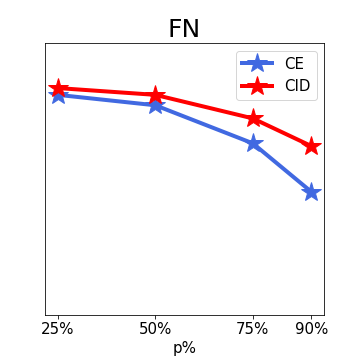}
\includegraphics[width = 0.19\linewidth]{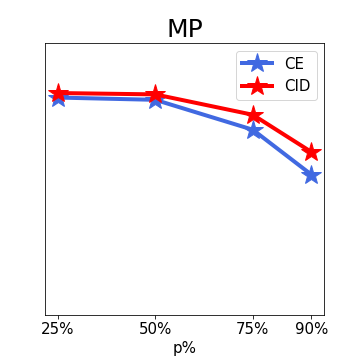}
\includegraphics[width = 0.19\linewidth]{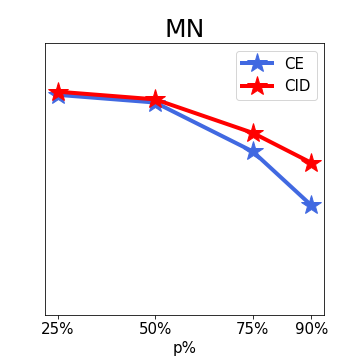}
\includegraphics[width = 0.19\linewidth]{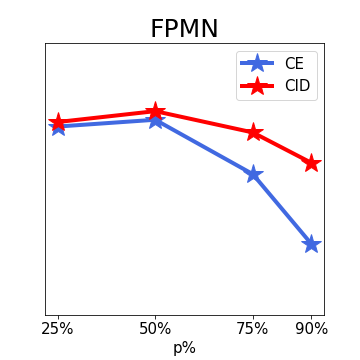}

\caption{Accuracy (Acc) and average per-Id Accuracy (Id Acc) results of different stress tests on the CelebA dataset for the smiling task label. It can be observed that the trend is similar to other metrics in terms of CID being more robust to bias and maintaining higher overall accuracy.}
\label{fig:acc_id_acc_smiling_stress_test}
\end{figure*}

Here we provide additional results to complete the ones provided in Section \ref{sec:stress_test}. In Table \ref{tab:smiling_FNMP}
we provide results on the biased dataset under the FNMP (excluding 90\% of female-negative and male-positive smiling examples) setup. Trends are similar to the experiments included in the main manuscript under Section \ref{sec:stress_test}. We also provide results on the other task of CelebA, namely, Mouth Slightly Open (MSO) in Figure \ref{fig:MSO_stress_test}. As mentioned in the main manuscript, all trends are similar to the smiling task, in terms of CID being consistently more accurate and less sensitive to bias, compared to CE. Also, the gap between the two models increases as bias in the dataset increases. In Table~\ref{tab:smiling_FP_MN_FPMN}, we provide results on the biased CelebA dataset under FP 90\%, MN 90\%, and FPMN 90\% setup. Trends are also consistently similar to the smiling tasks under section~\ref{sec:stress_test}.
In addition, we also add the trends in accuracy and ID-accuracy in the smiling task in Figure \ref{fig:acc_id_acc_smiling_stress_test}.

\begin{table}[htbp]
    \centering
    \caption{Stress testing the models for the smiling task on the CelebA dataset, by eliminating 90\% of the FNMP subpopulation (90\% of the female-negative samples, and 90\% of the male-positive samples) in the training/validation set. CID achieves the best results on the non-modified test set. This is consistent with other settings reported in Section \ref{sec:stress_test}.}
\resizebox{\linewidth}{!}{ 
    \begin{tabular}{c|c|c|c|c|c|c}
    \toprule
      {\bf FNMP }    & Acc & Id Acc & 10\% Id Acc & $\delta_{\text{Id}}$ & AUMM & AUD\\ \hline
CE      & 87.61   &   87.28    &   60.47        &      0.1252     &  0.2137  & \underline{0.1479 }\\
IFW     &   87.75   &    86.28       &       60.13        &     0.1256           & 0.2161   & 0.1533 \\
DRO     &  87.85   &     \underline{87.45}   &        
\underline{60.64} &     \underline{0.1240}         &   \underline{0.2122} & 0.1535 \\
IRM     &   87.77  &  86.23     &        60.29     &        0.1256      &  0.2145  & 0.1493  \\
ARL     &   \underline{88.10}  &  86.82    &   59.33         &        0.1285     &  0.2194  & 0.1539 \\\hline
CID     & \textbf{88.44}    &   \textbf{87.93}     & \textbf{61.54}            &  \textbf{0.1224}          & \textbf{0.2068}  &\textbf{0.1465}\\
\bottomrule
\end{tabular}
}
\label{tab:smiling_FNMP}
\end{table}

\begin{table}[htbp]
\caption{Stree testing on the models for the mouth slightly open (MSO) task on CelebA dataset, by eliminating 90\% FP, MN, FPMN of a subpopulation in the training/validation set. CID achieves the best results.}
\resizebox{\linewidth}{!}{ 
\begin{tabular}{c|c|c|c|c|c|c}
\toprule
\textbf{FP}  & Acc   & Id Acc & 10\% Id Acc    & $\delta_{\text{Id}}$  & AUMM   & AUD \\ \hline
CE  & 91.70 & 91.29  & 67.58 & 0.1061 & 0.1726 &  0.1349   \\
IFW & \underline{92.44}     &  \underline{92.09}      &  \underline{69.53}     &  \underline{0.1023}
   &    \underline{0.1612}     &    \underline{0.1300}  \\
DRO & 91.81 & 91.33  & 67.33 & 0.1059 & 0.1707 &0.1323\\
IRM & 91.71 & 91.25  & 67.78 & 0.1083 & 0.1729 & 0.1346    \\
ARL     &   91.94  &  91.38   &        67.13    &       0.1071     &  0.1735  & 0.1311 \\\hline
CID &  \textbf{92.67} & \textbf{92.16}  & \textbf{69.91} & \textbf{0.1000} & \textbf{0.1589} &  \textbf{0.1126}\\ 
\bottomrule
\end{tabular}
}

\resizebox{\linewidth}{!}{ 
\begin{tabular}{c|c|c|c|c|c|c}
\toprule

\textbf{MN} & Acc   & Id Acc &  10\% Id Acc    & $\delta_{\text{Id}}$  & AUMM   & AUD \\ \hline
CE  & 92.36 & 91.67  & 68.33 & 0.1011 & 0.1662 &    0.1200 \\ 
IFW & \underline{92.67} & \underline{92.22}  & \underline{71.17} & \underline{0.0971} & \underline{0.1563} &   0.1265 \\
DRO & 92.54 & 91.89  & 68.37 & 0.1054 & 0.1672 &   \underline{0.1166}  \\
IRM &   92.62    &  91.96      &  68.46     &   0.1048     &  0.1657      & 0.1192    \\
ARL     &   92.34  &   91.86   &       68.59    &        0.1053     &  0.1679  & 0.1213 \\ \hline
CID & \textbf{92.90} & \textbf{92.32}  & \textbf{71.87} & \textbf{0.0949} & \textbf{0.1560} &  \textbf{0.1161}  \\
\bottomrule
\end{tabular}
}
\resizebox{\linewidth}{!}{ 

\begin{tabular}{c|c|c|c|c|c|c}
\toprule
\textbf{FPMN} & Acc   & Id Acc & 10\% Id Acc    & $\delta_{\text{Id}}$  & AUMM   & AUD \\ \hline
CE   & 89.99 & 89.39  & 64.22 & 0.1238 & 0.2011 &  0.1410   \\
IFW  & 89.88 & 89.05  & 65.82 & 0.1275 & 0.2033 & 0.1424    \\
DRO  & \underline{90.02} & \underline{89.42}  & \underline{65.93} & 0.1211 & 0.1988 &   \underline{0.1336 } \\
IRM  & 89.99 & 89.35  & 65.41 & 0.1212 & \underline{0.1984} &   0.1429  \\
ARL     &   90.00  &  89.40    &        63.63    &       \underline{ 0.1175}   &  0.1987  & 0.1381 \\\hline
CID  & \textbf{90.47} &  \textbf{ 89.94}     &  \textbf{67.05} & \textbf{0.1113} &   \textbf{0.1883}    &  \textbf{0.1295} \\ \bottomrule  
\end{tabular}
}
\label{tab:smiling_FP_MN_FPMN}
\end{table}

\subsection{Ablation Studies on $\tau$}
\label{sec:ablation_tau}
\begin{figure*}[h!]
    \centering
 \includegraphics[width = 0.24\linewidth]{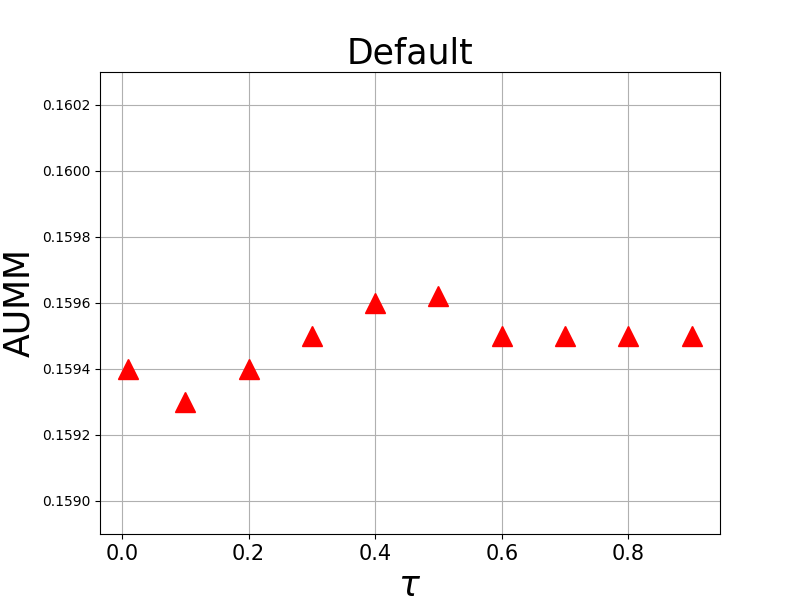}
  \includegraphics[width = 0.24\linewidth]{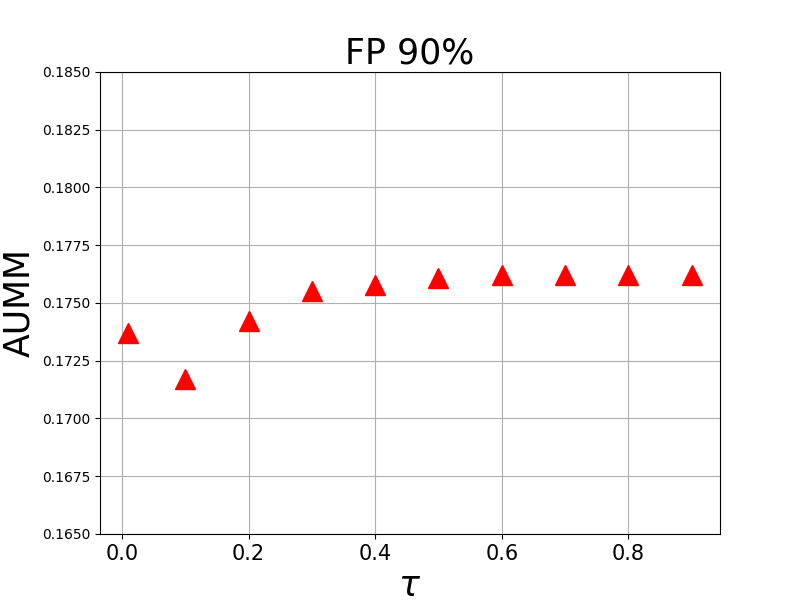}
     \includegraphics[width = 0.24\linewidth]{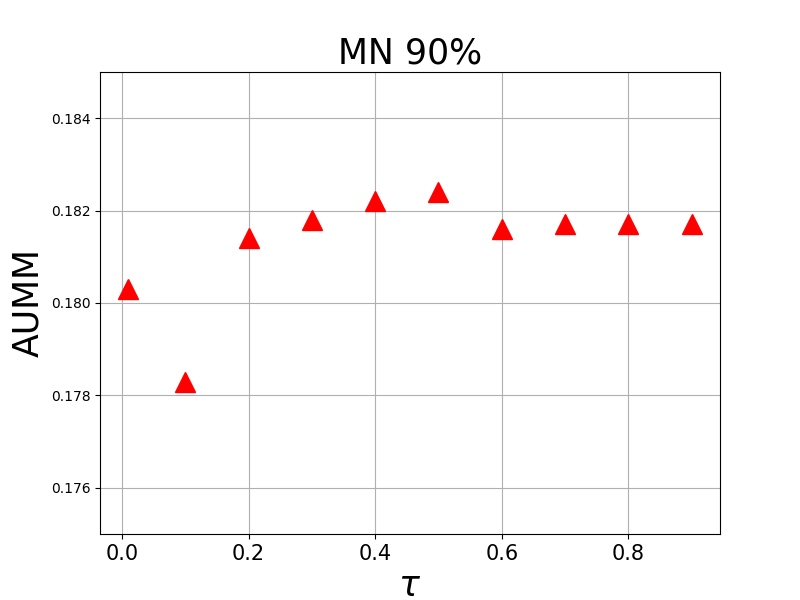}
         \includegraphics[width = 0.24\linewidth]{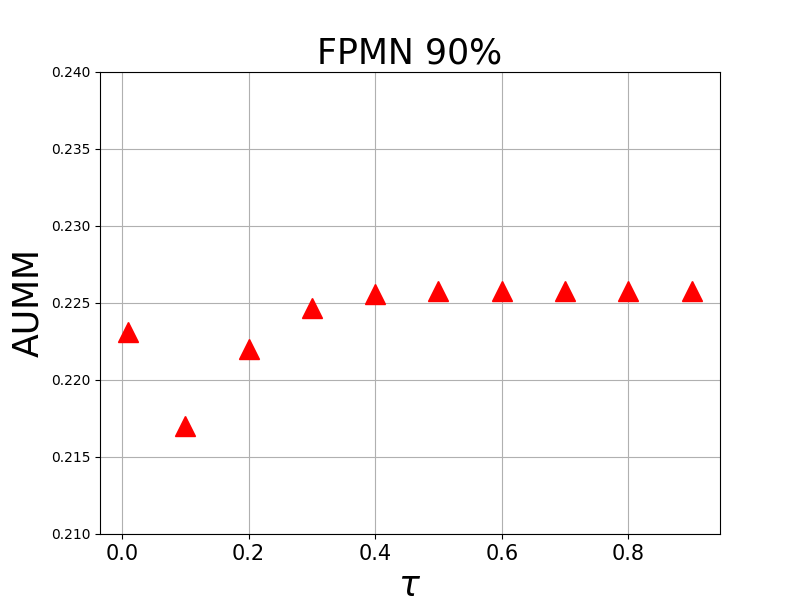}
         
  \includegraphics[width = 0.24\linewidth]{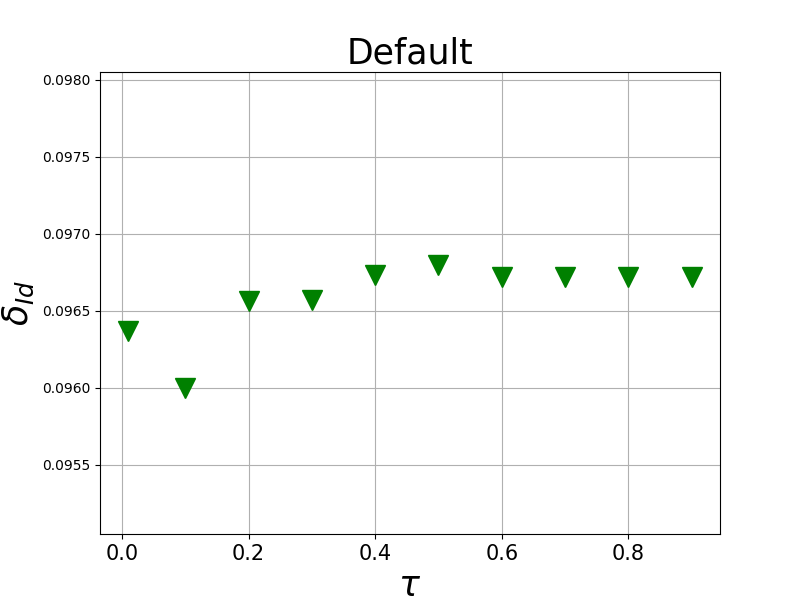}
  \includegraphics[width = 0.24\linewidth]{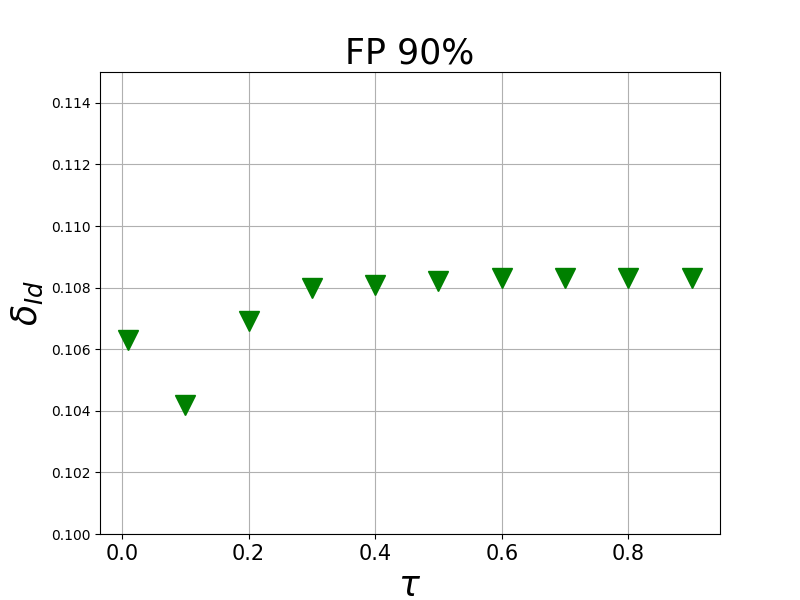}
  \includegraphics[width = 0.24\linewidth]{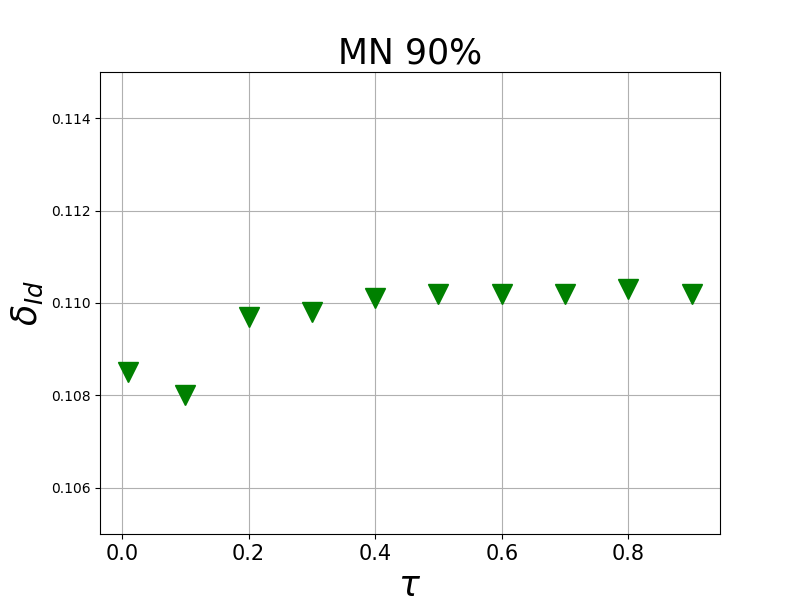}
  \includegraphics[width = 0.24\linewidth]{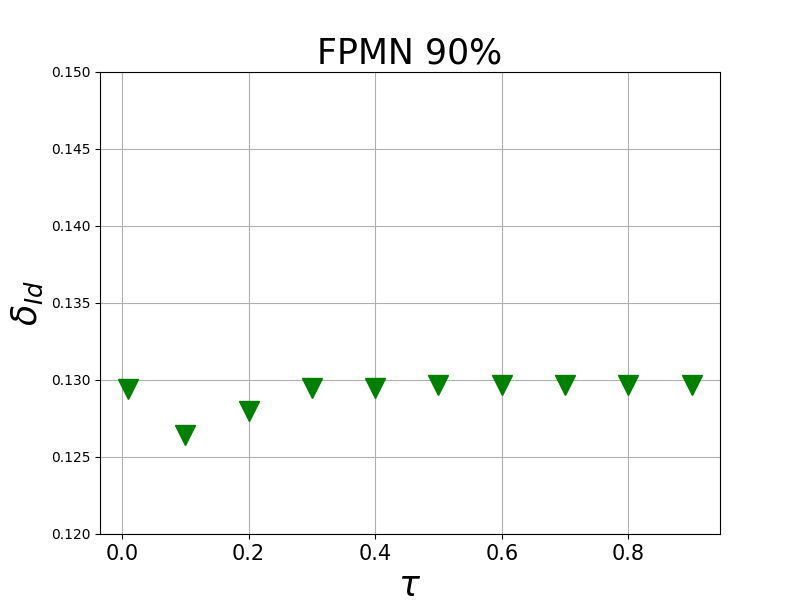}

   \includegraphics[width = 0.24\linewidth]{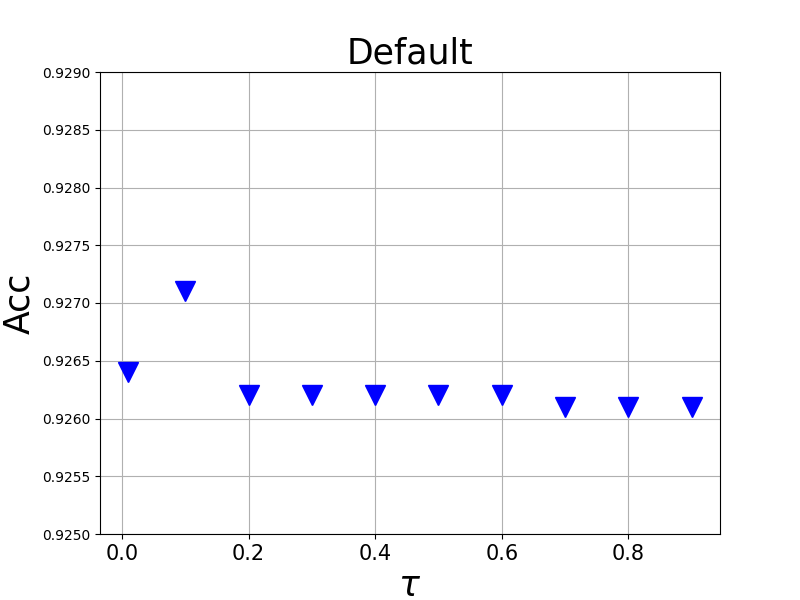}
  \includegraphics[width = 0.24\linewidth]{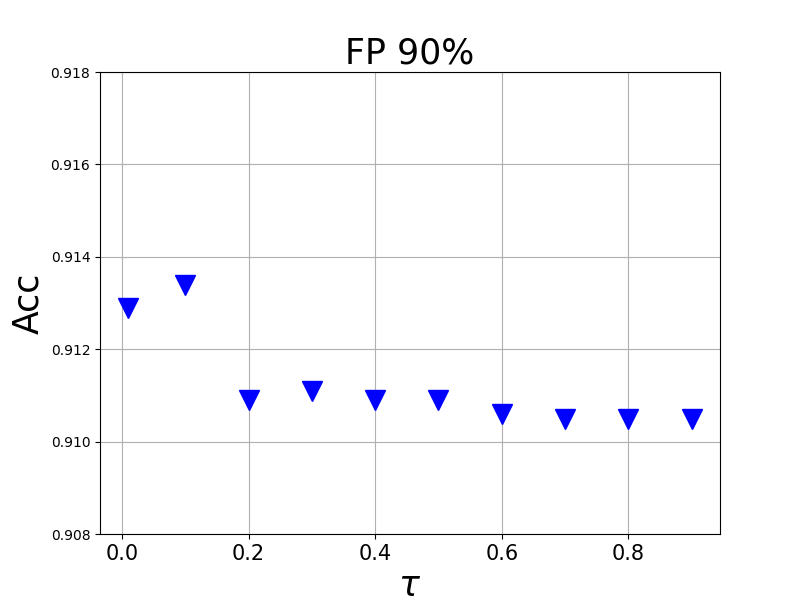}
  \includegraphics[width = 0.24\linewidth]{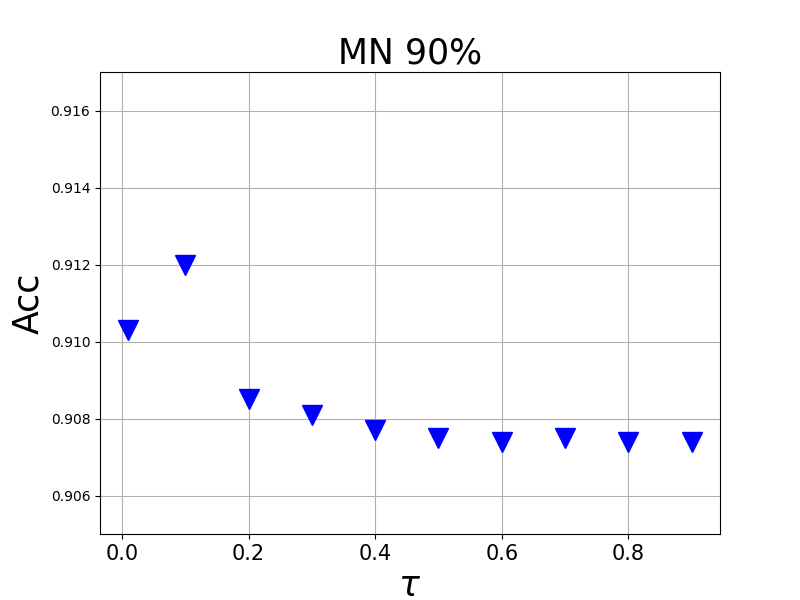}
  \includegraphics[width = 0.24\linewidth]{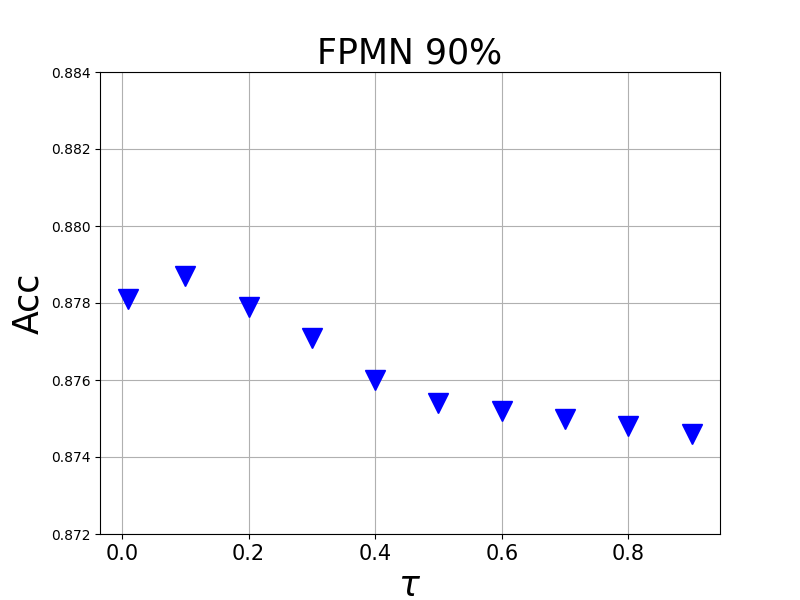}
\vspace{-0.1in}
    \caption{The robustness of the proposed CID methods in terms $\tau$.  As it can be observed maximum identity-robustness (minimum AUMM and $\delta_{\text{Id}}$), and maximum accuracy are achieved almost consistently at $\tau$=0.1}
    \label{fig:tau_ablation}
\end{figure*}

In this section, we provide ablation studies of $\tau$ on the smiling task on CelebA, to show its effect on the results of the proposed CID method. As mentioned in the main manuscript in Section \ref{sec:rpse}, and shown in Figure \ref{fig:soft_assignment_weights}, this hyper-parameter implicitly captures the neighborhood radius across which performance is encouraged to remain uniform. 
In Figure \ref{fig:tau_ablation}, we report the AUMM, $\delta_{\text{Id}}$, and Accuracy by varying $\tau \in [0.01, 0.9]$ on the default dataset and stressed FP 90\%, MN 90\%, FPMN 90\% datasets. As it can be observed we can see the optimal value (lowest AUMM) is always achieved at 0.1. We hypothesize that as this maps to a specific radius in the embedding space, it captures the optimal neighborhood size for an individual person in the proxy embedding space. This is fairly consistent with the results reported in \cite{ardeshir2022estimating} regarding a specific radius being best for estimating individual fairness.

\subsection{Rawlsian Min-Max (RMM) on gender and age}
\label{sec:Aba_RMM}
To complete our experiments, we report fairness in terms of identity-related gender and age attributes.  Given that identity groups are defined by a unique set of characteristics, enhancing the fairness of identity can result in improved fairness across gender and age groups.
As gender ($Male$ or not) and age ($Young$ or not) are labeled as binary in CelebA, we evaluate Rawlsian max-min (RMM) to measure disparity.
 We compare CID with other baselines and report the results in Table~\ref{tab:gender_age_rmm} under different setups on the Smiling task of CelebA, the improved (reduced) RMM verifies the effectiveness of the CID method on binary identity-related attributes.


\begin{table}[h!]
\centering
\caption{Gender and age Rawlsian min-max results on smiling tasks}
\label{tab:gender_age_rmm}
\resizebox{0.4\textwidth}{!}{%
\begin{tabular}{c|c|c
|c|c}\toprule
  Gender  & Default & MP 90\% & MN 90\% & FPMN 90\% \\ \toprule
CE  &    0.01717     &   0.05047   &    0.07859  &   0.05940     \\
IFW &    0.01748     &  \underline{ 0.04116 }  &   \underline{0.05415}   &   0.07738     \\
DRO &      0.01671   &  0.04901    &   0.07635   &   0.05933     \\
IRM &     \underline{0.01645}   &   0.05117   &    0.07792   &  \underline{0.05722}      \\ 
ARL &    0.01678   &  0.05366   &   0.07785  & 0.05910     \\\hline
CID &   \textbf{ 0.01633}    &  \textbf{0.03553}    & \textbf{0.04893}     &    \textbf{0.04428}  \\
\bottomrule
\end{tabular}%
}

\resizebox{0.4\textwidth}{!}{%
\begin{tabular}{c|c|c
|c|c}\toprule
  Age  & Default & MP 90\% & MN 90\% & FPMN 90\% \\ \toprule
CE  &    0.02251     &  0.04167    &  0.06421    &   0.07548     \\
IFW &    0.02289     & \underline{0.03378}    &  \underline{0.05536}   &    0.08442  \\
DRO &     0.02336  &  0.03838    &  0.06258    &   \underline{0.07518}     \\
IRM &      \underline{0.02251} &   0.04001   &  0.06556   &   0.07698     \\ 
ARL &    0.02271   &  0.04136   &   0.06744  &0.07624\\  \hline
CID &    \textbf{0.02231}    &    \textbf{0.03131}    & \textbf{0.05288}     &   \textbf{0.06411} \\
\bottomrule
\end{tabular}%
}
\end{table}

\subsection{Connects and Comparison with other baselines}
\label{sec:cc_baselines}

In this section, we draw parallels and provide comparisons of CID with the methods that can be broadly categorized into the sample weighting scheme including ARL, DRO, and IFW class-balancing method.

\subsubsection{Adversarially Reweighted Learning (ARL)}
 ARL~\cite{lahoti2020fairness} is a fairness-aware method that designed to handle fairness under awareness challenges. To achieve this, ARL makes use of adversarial training to emphasize more on the samples 
 that make significant errors and improve improve accuracy for the protected groups. Specifically, ARL optimizes the following min-max objective:
 \begin{align}
 \label{eqn:appendix_ARL}
    \min_\w\max_{\phi}  \lambda_{\phi}(\x_i, y_i) \ell_i(\w;\x_i,y_i)
\end{align}
where $\w$ is the model (learner) parameters, and $\phi$ denotes the adversarial network, $f_\phi$ represents the output of adversarial network
and $\lambda_{\phi}(\x_i,y_i) = 1/n + \frac{f_{\phi}(\x_i,y_i)}{\sum_{i=1}^n f_{\phi}(\x_i,y_i)}$ is the adversarial assignment of the weight vector $\lambda_\phi: f_{\phi} \rightarrow R$ so as to maximize the expected loss and $f_\phi: X\times Y\rightarrow R$ is the output of the adversary network designed to identify the regions where the model makes significant errors.

In our case, we aim to balance the positive and negative samples for each identity during the training and emphasize more on the minorities to achieve fairness under unawareness. 
 As the same facial features close to each other in the face recognition space, we make use of face recognition embeddings as proxy vectors of identities due to the unavailable of ground truth label information. We assign conditional inverse density robust weight $p^\tau_i$ to each sample defined in the neighborhood of the proxy space to balance the positive and negative samples that share the same facial features, i.e, identity.
\begin{align}
    \min_\w \sum\limits_{i=1}^n\frac{p^\tau_i}{Z_{y_i}} \ell_i(\w;\x_i,y_i)
\end{align}
where $n$ represents the size of $\D$, i.e, $n = |\D|$, $  p_{i}^\tau =  \exp(\frac{\z_i^\top\z_i}{\tau})/ \sum_{k=1}^{|\D_{y_i}|}\exp(\frac{\z_i^\top\z_k}{\tau})$  and $\tau$ represents the size of the neighborhood within which samples are considered to be from the same identity group. $Z_{y_i} = \sum_{j\in \D_{y_i}}p_j^{\tau}$ to balance class contributions.

Hence by replace $\lambda_\phi(\x_i,y_i)$ with $p_i^\tau$ in Euqation~(\ref{eqn:appendix_ARL}), ARL share the same objective with our proposed method.
Compare with $p_i^\tau$, $\lambda_\phi(\x_i,y_i)$ is obtained via maximizing the objectives combined with the learner prediction losses, i.e, $\max_{\phi} \lambda_{\phi}(\x_i, y_i) \ell_i(\w;\x_i,y_i)$.
Hence, the $\lambda_{\phi}(\x_i, y_i)$ is obtained based on the sample loss predictions rather than features that represent group information such as identity, which leads to poorer performance of ARL than CID in terms of identity fairness, as observed in all experiments

\noindent\textbf{ARL Implementation} 
The same as other baselines, the learner is implemented using ResNet18. For the adversarial head, we implement a linear model with sigmoid activation function where the input is the representation feature learned from learner. The learning rate of adversarial learner $\phi$ is tuned in $\{1e$-$2, 1e$-$3, 1e$-$4\}$.

 \subsubsection{Distributionally Robust Learning}
DRO~\cite{li2021tilted, qi2020attentional} has been show an effective methods to improve the worst group accuracy such that the fairness can be improved.
Similar to ARL, DRO 

\begin{align}
    \min\limits_{\w}\max\limits_{\p\in\U} p_i\ell_i(\w;\x_i,y_i)
\end{align}
in which $\p$ is defined in the uncertainty set $\U$.
Hence, after getting $\p^*$ by solving the inner maximization problem, the DRO reduces to a instance sample weighting method that
\begin{align}
    \min\limits_{\w} p^*_i\ell_i(\w;\x_i,y_i)
\end{align}
To be more specific, when $\U = \{\p\in\Delta,   \sum\limits_{i=1}^n p_i\log n p_i \leq \rho\}$ in ~\cite{li2021tilted, qi2020attentional}, where $\p\in\Delta:\sum\limits_{i=1}^n p_i = 1$, $\sum\limits_{i=1}^n p_i\log n p_i \leq\rho$ represents the KL-divergence discrepancy constraint between $\p$ and uniform sampling weights $\frac{\textbf{1}}{n}, 1\in\R^n$. Given this uncertainty set,  $p_i^*$ has the close form of
$p_i^* = \exp(\frac{\ell_i(\w;\x_i,y_i)}{\nu})/\sum\limits_{i=1}^{n}\exp(\frac{\ell_i(\w;\x_i,y_i)}{\nu})$, where $\nu$ is a Lagrange multiplier introduced by $\U$.

Therefore, we can see the instance weights $p_i^*$ for each sample introduced by DRO is proportional to the loss scales, while our fairness robust weights $p_i^{\tau}$ defined on the pairwise similarity in the neighborhood face proxy embeddings of sample $(\x_i,y_i)$ to balance the contribution of positive and negative classes for any arbitrary facial features that are considered belonging to the same identity.
 \subsubsection{Inverse Frequency Weighting (IFW)}

IFW~\cite{huang2016learning,  wang2017learning} aims to balance the contributions of different classes by assigning class-balancing weights to each sample that is proportional to the number of samples for each class, i.e,
\begin{align}
    \sum\limits_{i=1}^n p_i\ell_{i}(\w;\x_i,y_i)
\end{align}
where $p_i \propto \frac{1}{N_{y_i}}$, where $N_{y_i}$ represents the frequency of class $y_i$.
As we explained in Section~\ref{sec:rpse}, our assign robust weights $p_i^\tau\rightarrow \frac{1}{|\D_{y_i}|}$ when $\tau \rightarrow \infty$ such that CID reduces to IFW.
Hence, IFW is a special case of CID.  When $\tau \in (0,\infty)$, CID considers more fine-grained rarity of local neighborhood face proxy embedding, i.e, $p_i^\tau $ balances the positive and negative samples within each local neighborhood which serves as a proxy of identity, to improve identity robustness.

\end{document}